\documentclass[sigconf]{acmart}
\AtBeginDocument{%
  \providecommand\BibTeX{{%
    \normalfont B\kern-0.5em{\scshape i\kern-0.25em b}\kern-0.8em\TeX}}}

\usepackage[utf8]{inputenc} 
\usepackage[T1]{fontenc}    
\usepackage{hyperref}       
\usepackage{url}            
\usepackage{booktabs}       
\usepackage{amsfonts}       
\usepackage{nicefrac}       
\usepackage{microtype}      
\usepackage{xcolor}         
\usepackage{multirow}

\usepackage{caption} 
\usepackage{subfigure}
\usepackage{pgfplots}
\usepackage{booktabs} 
\usepackage{amsmath}
\usepackage{amsthm}
\usepackage{pgfplots}
\usetikzlibrary{matrix}
\usepgfplotslibrary{groupplots}
\usetikzlibrary{patterns}
\usepackage{algorithm}
\usepackage{algorithmic}
\usepackage{colortbl}
\usepackage{tabu}
\usepackage{url}
 \usepackage{balance}
\usepackage{threeparttable}
\usepackage{natbib}
\setcitestyle{numbers}
\setcitestyle{square}
\newtheorem{innercustomgeneric}{\customgenericname}
\providecommand{\customgenericname}{}
\newcommand{\newcustomtheorem}[2]{%
  \newenvironment{#1}[1]
  {%
   \renewcommand\customgenericname{#2}%
   \renewcommand\theinnercustomgeneric{##1}%
   \innercustomgeneric
  }

  {\endinnercustomgeneric}
}
\newcustomtheorem{customthm}{Theorem}
\newcustomtheorem{customlemma}{Lemma}
\usepackage{filecontents}

\usepackage{graphicx}

\usepackage{pgfplots}
\pgfplotsset{width=5cm,height=4cm}
\usetikzlibrary{matrix}
\usepgfplotslibrary{external}

\usepackage{pgfplots}
\usepgfplotslibrary{groupplots}
\pgfplotsset{compat=1.17}

\copyrightyear{2023} 
\acmYear{2023} 
\setcopyright{rightsretained} 
\acmConference[CIKM '23]{Proceedings of the 32nd ACM International
Conference on Information and Knowledge Management}{October 21--25,
2023}{Birmingham, United Kingdom}
\acmBooktitle{Proceedings of the 32nd ACM International Conference on
Information and Knowledge Management (CIKM '23), October 21--25, 2023,
Birmingham, United Kingdom}\acmDOI{10.1145/3583780.3614955}
\acmISBN{979-8-4007-0124-5/23/10}
\begin{document}

\title{Low-bit Quantization for Deep Graph Neural Networks with Smoothness-aware Message Propagation}




\author{Shuang Wang}
\orcid{0000-0002-2498-4740}
\affiliation{%
  \institution{University of Warwick}
  \city{Coventry}
  \country{United Kingdom}}
\email{Shuang.Wang.1@warwick.ac.uk}

\author{Bahaeddin Eravci}
\orcid{0000-0002-9006-917X}
\affiliation{%
  \institution{TOBB University of Economics and Technology}
  \city{Ankara}
  \country{Turkey}
  }
\email{beravci@gmail.com}

\author{Rustam Guliyev}
\orcid{0009-0008-5515-0766}
\affiliation{%
  \institution{University of Warwick}
  \city{Coventry}
  \country{United Kingdom}}
\email{Rustam.Guliyev@warwick.ac.uk}

\author{Hakan Ferhatosmanoglu}
\orcid{0000-0002-5181-4712}
\affiliation{%
  \institution{University of Warwick}
  \city{Coventry}
  \country{United Kingdom}
}
\email{Hakan.F@warwick.ac.uk}
\authornote{Also with Amazon Web Services. This publication presents work performed at the University of Warwick and is not associated with Amazon.}



\begin{abstract} 

Graph Neural Network (GNN) training and inference involve significant challenges of scalability with respect to both model sizes and number of layers, resulting in degradation of efficiency and accuracy for large and deep GNNs. We present an end-to-end solution that aims to address these challenges for efficient GNNs in resource constrained environments while avoiding the oversmoothing problem in deep GNNs. We introduce a quantization based approach for all stages of GNNs, from message passing in training to node classification, compressing the model and enabling efficient processing. The proposed GNN quantizer learns quantization ranges and reduces the model size with comparable accuracy even under low-bit quantization. To scale with the number of layers, we devise a message propagation mechanism in training that controls layer-wise changes of similarities between neighboring nodes. This objective is incorporated into a Lagrangian function with constraints and a differential multiplier method is utilized to iteratively find optimal embeddings. This mitigates  oversmoothing and suppresses the quantization error to a bound.  
Significant improvements are demonstrated over state-of-the-art quantization methods and deep GNN approaches in both full-precision and quantized models. The proposed quantizer demonstrates superior performance in INT2 configurations across all stages of GNN, achieving a notable level of accuracy. In contrast, existing quantization approaches fail to generate satisfactory accuracy levels. Finally, the inference with INT2 and INT4 representations exhibits a speedup of 5.11 $\times$ and 4.70 $\times$ compared to full precision counterparts, respectively. 
 
\end{abstract}

\maketitle

\section{Introduction}

Data analytics and machine learning on large graphs encompass a wide array of applications, including recommender systems, social networks, web analysis, and computational biochemistry. Some tasks within this scope include node classification, community detection, link prediction, 
reachability analysis, and influence optimization.
Recently, Graph Neural Networks (GNNs) have shown to be  effective for learning over graphs~\cite{TNNLS2021GNNSurvey}. 
GNNs utilize an iterative process, aggregating features from neighboring nodes through learnable parameters, thus generating rich and informative embeddings.


The versatility of GNNs often comes at a price – elevated memory and computation demands. This poses challenges when scaling up to larger graphs and deeper models.
Large-scale graphs naturally increase the storage costs and the neighborhood size during the aggregation phase. 
While deeper models, with more iterative layers, add computational strain, they do capture intricate relationships by broadening the nodes' receptive fields. 
To counter these, recently, quantization approaches were developed to compress both the model and graph, aiming to reduce storage, computation, and power requirements for inference workloads~\cite{zhu2023rm, tailor2020degree, ding2021vq, feng2020sgquant}.

Quantization is the process of mapping continuous numerical values into 
smaller sized representations (e.g., using 8-bits). There are a variety of  quantization methods developed for data-intensive tasks, ranging from multi-dimensional indexing for range and similarity queries~\cite{IndexQuantization,VAplusquantization2000,tuncel2002vq,wang2020deltapq,wang2020ppq,highdimMLindex2021,zhan2021jointly} to processing convolutional and recurrent neural networks~\cite{lin2017towards,yao2020hawqv3, wang2018hitnet}.
For GNNs, quantization is useful in many practical settings, such as resource-efficient representation learning, reducing energy and communication in sensors, IoT and mobile devices, on-device and embedded learning, managing data/models in distributed and edge computing, and recommender systems which commonly employ graph neural networks.

Unlike conventional applications of quantization, GNNs present unique challenges due to their intrinsic characteristics, which are not, effectively, addressed by the current methods.
(i) The process of neighborhood aggregation in GNNs can lead to significant variance in high in-degree node embeddings,
thereby exacerbating the quantization error, especially in low-bit cases~\cite{tailor2020degree}.
(ii) As GNNs deepen, they tend to experience the "oversmoothing" issue where each embedding loses its discriminative information due to the repeated, unregulated message passing~\cite{klicpera2018predict}. It is important to understand if this problem remains or is aggravated with the introduction of model quantization. Thus, while reducing GNN size and enabling compressed processing are pivotal for performance efficiency, addressing oversmoothing is crucial to ensure accuracy, especially in deeper models. 


While recent studies~\cite{tailor2020degree, zhu2023rm,huang2022epquant} have delved into GNN quantization, the problem is far from being solved and there is no effective solution for low-bit quantization that scales for deeper GNNs. Our paper underscores this challenge, revealing that state-of-the-art GNN quantization methods undergo significant degradation at low bit counts (INT4 and INT2).
This is more pronounced in deeper GNNs, due to accumulated layer-by-layer quantization errors. We aim to address these intricacies and develop an end-to-end solution. 



Our solution involves a quantizer that learns the quantization ranges (QLR) along with a skewness-aware bitwise truncation (BT$^{*}$) mechanism. Additionally, we introduce a smoothness-aware message propagation scheme (SMP) to counter the oversmoothing issue in quantized models. This quantizer determines an optimized, data-aware learnable range grounded in the input data distribution, thereby minimizing model redundancy. It is shown to retain its effectiveness with low-bit representations, which makes it apt for large deep GNNs. The skewness-aware truncation embedded within the quantizer improves the accuracy particularly in low-bit (INT2) scenarios.
Our message propagation scheme aims to mitigate oversmoothing in deep GNNs by constraining the layer-wise shifts in similarities among neighboring nodes.
Furthermore, we prove that by using SMP, the quantization error can be suppressed to a bound. 
Finally, we demonstrate the efficiency and accuracy of our solution through node classification accuracy on quantized GNN models.

Experimental results demonstrate improvements over the state-of-the-art approaches across various performance measures and workloads. Specifically, our quantizer (QLR) demonstrates remarkable advancements in low-bit quantization, 
outperforming existing quantization methods while resulting in reduced model sizes. For deeper GNNs, our SMP method delivers more accurate classification compared to other deep GNN approaches both in full-precision and quantized versions.
The low-bit quantized SMP, using QLR, achieves greater improvement over alternative deep quantized GNN approaches with the help of the quantization error bound with SMP.
BT$^{*}$ improves node classification accuracy on large datasets with INT2 representation, making it comparable to INT8 accuracy. 
We also show that the
INT2 quantization model can yield an inference speedup of 5.11 $\times$ compared to the full-precision model.

\vspace{-4.0mm}
\section{Related Work}\label{relate_work}
Quantization has been commonly employed for neural network (NN) models~\cite{gholami2021NNQuantizationsurvey}. NN training is bottlenecked by high memory requirements to handle large data involving intermediate results and feature maps~\citep{band2020memflow}.  
NNs can be trained with low precision using dynamic fixed point arithmetic~\cite{courbariaux2015trainingLowPrecision}. 

Quantization for neural networks can be performed during or after training. The post-training approaches quantize weights or activation of neural networks on a pre-trained full-precision model~\cite{han2015deep,banner2018post}. Their low-bit quantization performance incurs significant accuracy degradation. The quantization-aware training aims to avoid this performance degradation~\cite{bengio2013estimating, dong2019hawq}. A useful technique is to expose errors from the quantization operation to the forward pass at model training and use straight-through estimator(STE) to compute the gradients~\cite{bengio2013estimating}. \citet{Ron2018Scalable8BitTraining} provide a theoretical analysis showing considerable room for quantization under Gaussian weight assumption leading to 8-bit DNNs with comparable accuracy. The success of quantization has led to binary NNs (BNN) drastically reducing computation and memory requirements using hardware-supported bitwise operations with strong precision performances~\cite{Hubara2017QuantizedNN}. 
The efficacy of high-order bit representations, involving bitwise truncation applied to 32-bit word embeddings, has been demonstrated in previous studies~\cite{chen2013expressive}.


Pioneering work on learning node representations~\cite{Scarselli2009GNN} has been followed by variants of architectures, utilizing convolutions~\citep{henaff2015deepGCN,niepert2016LearningGCN,kipf2017GCN} and autoencoder structures~\cite{Cao2016GnnAutoencoder,kipf2016variational}.
GNN based representations have been used for various analytics tasks, including  node similarity search~\cite{echihabi2020high}, link prediction~\cite{rossi2022explaining}, and entity disambiguation~\cite{vretinaris2021medical}. Solutions for scaling GNNs mostly focus on distributed processing~\citep{zhang13agl,demirci2022scalable,zheng2022bytegnn}. Scalability challenges on large graphs have also been studied in the context of 
memory optimizations~\cite{shao2013trinity} and scalable processing~\cite{sahu2017ubiquity, mondal2012managing, sahu2020LargeGraphSurvey}.


GNN quantization have started to receive attention in recent years. \citet{tailor2020degree} propose quantization-aware training for GNNs, where high in-degree nodes are selected for full-precision training while all other nodes are converted to INT8/INT4. This can achieve reasonable accuracy especially on INT8 models. \citet{huang2022epquant} employ product quantization to compress input data but do not address the more challenging task of quantization of parameters.
A recent GNN quantization approach~\cite{zhu2023rm} addresses low-bit representation of the weights and input features by learning the parameters that are equal with the weight dimension and the number of input nodes, respectively, while leaving the core message propagation unquantized. However, this approach necessitates the learning of parameters that scale proportionally with the number of input nodes, resulting in considerable storage and space overheads.
Neural Architecture Search (NAS) is used to span possible quantization levels suggesting an INT4 weight and INT8 activation as an effective strategy for GNNs~\cite{zhao2020learned}. Recent studies adapt binary NN methods for GNNs~\cite{Wang21BinarizedGNN,Mehdi20BinaryGNN} offering a trade-off between time/space efficiency and classification accuracy. These methods typically either need an additional teacher model for knowledge distillation or learn binary weights for each layer's input message, which require higher storage and computational load than a typical quantization based approach.

Towards addressing oversmoothing in deep GNNs, \citet{liu2021elastic} propose Elastic Graph Neural Network with long-range information propagation using  $\ell_{1}$ and $\ell_{2}$-based graph smoothing. APPNP~\cite{klicpera2018predict} addresses the oversmoothing with a propagation scheme based on an approximation of personalized PageRank. \citet{zhu2021interpreting} proposed low-pass and high-pass filtering kernels which have empirically reduced the effect of oversmoothing. DropEdge~\cite{rong2019dropedge}  aims to address the oversmoothing by dropping a number of edges, which can be interpreted as both a data augmentation method generating random deformed graphs and message passing reducer by sparsifying edge connections. PairNorm~\cite{zhao2019pairnorm} quantifies the oversmoothing and proposes a two-step
center-and-scale normalization layer to prevent nodes converging to similar representations. Compared to enforcing local smoothness, our method, constrains the layer-wise message propagation to counteract oversmoothing, which achieves performance improvements over the prior approaches as also demonstrated in our experiments.


\section{Preliminaries and Analysis}
We first provide the technical background, covering quantization for GNNs, analysis of quantization errors, and the oversmoothing problem in GNNs. 


\subsection{GNN Basics}
A graph  $\mathcal{G}$ is represented as $\mathcal{G}$ = ($V$, $E$, $\mathbf{X}$), where $V = \{v_1, \cdots , v_n\}$ is the set of $n$ nodes $|V|=n$, $E$ is the set of all edges, $\mathbf{H^l}$ = $[\mathbf{h}^l_{1}, \cdots, \mathbf{h}^l_{n}]^{\top}$ is the node feature (embedding) matrix for layer $l \in L$ where $L$ represents the number of layers in $\mathcal{G}$, and $\mathbf{h}_{i} \in \mathbb{R}^{d_{l}}$ is the feature vector for $v_i\in V$ node with initial $\mathbf{H^0}=\mathbf{X}$.  The adjacency matrix of $\mathcal{G}$ is a binary matrix $\mathbf{A} \in \mathbb{R}^{n \times n}$, where $\mathbf{A}(i,j)=1$ if the edge between nodes $v_i$ and $v_j$ exists ($e_{ij} \in E$), and $0$ otherwise.

GNNs comprise a sequence of layers with three main functions for each layer: message, aggregate and update. This framework is generally called Message Passing NNs (MPNN)~\cite{Gilmer2017MPNN}. Each message, which is a flow of data from nodes' neighbors, is aggregated and joined with existing embedding to form a new one for the respective node as given in Equation~\ref{eq:MPNN}, where $\mathcal{N}_u$ are the neighboring nodes of node $u$. The feed-forward iteration starts with the initial embeddings, $\mathbf{h}^0_u$=$\mathbf{x}_{u}$.
\begin{equation}
\mathbf{h}^{(l+1)}_{u} ={Update}_l(\mathbf{h}^l_u,Aggregate_l({\mathbf{h}_v^l, \forall v \in \mathcal{N}_u})
\label{eq:MPNN}
\end{equation}

Various GNN architectures are proposed in the literature, essentially, varying the message, aggregate and update functions~\cite{TNNLS2021GNNSurvey}. We consider, the popular, GCN (Graph Convolutional Network) architecture where the update function given in Equation~\ref{eq:GCN} with activation function $\sigma$ and learnable weight matrix $\mathbf{W}_l$~\cite{kipf2017GCN}.
\begin{equation}
\mathbf{h}_{(l+1)}^{u} =\sigma \bigg( \sum_{v \in \mathcal{N}^u \cup u}\frac{1}{\sqrt{d_u d_v}}\mathbf{W}_l\mathbf{h}_{l}^{v}) \bigg)
\label{eq:GCN}
\end{equation}%

Recently, a different perspective on common GNN models was proposed by \citet{ma2021unified}, where the authors unified different GNN models, such as GCN, GAT, PPNP, and APPNP, by posing them as solutions to the graph denoising problem.
\begin{equation}
\centering
\begin{aligned}
&\underset{\mathbf{H}}{\text{argmin}} \quad \frac{1}{2}\|{\mathbf{H-X}}\|_{F}^{2} + \frac{\mu}{2} \mathbf{tr(H^{\top}LH)} 
\end{aligned}
\label{eq:basic_denoising}
\end{equation}
where $\mathbf{H}$=$[\mathbf{h}_{1}, \cdots, \mathbf{h}_{n}]^{\top}$$\in$ $\mathbb{R}^{n\times d}$, $\mathbf{L}$$\in$$\mathbb{R}^{n\times n}$ represent final representations and the graph Laplacian matrix respectively. The first term drives the embeddings $\mathbf{H}$ closer to the graph input features $\mathbf{X}$ while the second term imposes global smoothness by enforcing similarity amongst the connected nodes, as
\begin{equation}
\mathbf{tr(H^{\top}LH)} =\sum_{(v_{i},v_{j}) \in E}\| {\mathbf{h}_{i}} - {\mathbf{h}_{j}}\|_{2}^{2}
\label{eq:edgedifference}
\vspace{-1mm}
\end{equation}
Following this, \textbf{EMP} (Elastic Message Passing)~\cite{liu2021elastic} method was proposed enabling $\ell_{1}$ based smoothing constraints on GNNs. 



\vspace{-3pt}
\subsection{Challenges with Deeper GNNs}
While in traditional ML, deeper models can extract more powerful representations, for GNNs this inherently leads to several major challenges.
First, as the depth increases, GNNs demand exponentially more computations and larger storage to be managed and processed, which makes their deployment on resource constrained platforms more challenging. We seek to design an inference-friendly quantizer, i.e., performing inference directly on quantized elements with high accuracy.
Second, deeper GNNs suffer from the oversmoothing problem, where node representations converge to indistinguishable embeddings, degrading accuracy of downstream tasks. It was shown that GCN exponentially loses its expressive power for node classification tasks in many practical cases~\cite{oono2019GraphNeuralNetworks}.


There are some proposals towards mitigating the oversmoothing problem for full precision models~\cite{klicpera2018predict,liu2021elastic,rong2019dropedge,zhao2019pairnorm,liu2020towards} as discussed in Section \ref{relate_work}, including \textbf{DropEdge}, \textbf{PairNorm}, \textbf{APPNP} and \textbf{EMP}. 
Our experiments confirm that \textbf{DropEdge} and \textbf{PairNorm} 
are particularly ineffective for low-bit quantization. 
These methods do not consider the smoothness of message propagation amongst layers, resulting in accuracy drops and unrestricted quantization error especially in low-bit cases. In contrast, we seek layer-wise smoothness by enforcing constraints at message propagation, restricting the quantization error, and denoising the message passing procedure which lead to enhanced accuracy in low-bit quantization.  

\subsection{Quantization Basics}\label{quantization_error_analysis}
Quantization is the process of mapping continuous data, e.g., parameters, weights and activations of neural networks, to smaller sized representations.
In the scope of our analysis, we denote $U$ (e.g. $\mathbf{H^{l}}$)  as a high-precision 
tensor-valued random variable with probability density function $f_U(u)$.
A tensor is commonly quantized between its maximum and minimum \emph{observed} values~\citep{jacob2017gemmlowp}.
Considering observed values as $U_{o}$ $\in$ $[\alpha, \beta]$ and the corresponding $b$-bit quantized values as $U_q$ $\in$ $[\alpha_q, \beta_q]$, the quantization function is given by
 \vspace{-1mm}
\begin{equation}
Q(U, s, z) = \mathrm{clip}\Big(\lfloor\frac{U}{s}+z\rceil, \alpha_q, \beta_q\Big) = U_q
\label{eq:quantization_function}
\end{equation}
\vspace{-1mm}
where $s$=$\frac{\beta-\alpha}{\beta_q-\alpha_q}$ is the scale, $\lfloor \cdot \rceil$ denotes the round function, 
and $z$=$\lfloor\frac{\beta\alpha_q - \alpha\beta_q}{\beta-\alpha}\rceil$ is the zero point. The corresponding de-quantization function is as follows
\vspace{-2mm}
\begin{equation}
D(U_q, s, z) =s(U_q-z)
\label{eq:dequantization_function}
\end{equation}

The range $[\alpha, \beta]$ is usually partitioned into 2$^{b}$ equal interval regions with a quantization step $\Delta$=$\frac{\beta - \alpha}{2^{b}}$.
Given that de-quantized value in Equation~\ref{eq:dequantization_function} is represented as $\hat{U}$, the mean squared error (MSE) 
between $U$ and $\hat{U}$ 
is given by
\begin{equation}
\small
\thinmuskip=1mu
\medmuskip=2mu
\begin{aligned}
\small
E[(U - \hat{U})^{2}] =
{\int_{-\infty}^{\alpha}f_U(u)(u-\hat{\alpha})^2 du}
+ 
 {\int_{\beta}^{\infty}f_U(u)(u-\hat{\beta})^2du}
\\
\hspace{-8mm}+{\sum_{i=0}^{2^b-1}\int_{\alpha+i\Delta}^{\alpha+(i+1)\Delta}f_U(u)(u-\hat{u})^2du}
\end{aligned}
\label{eq:quantized_MSE}
\vspace{-0.1cm}
\end{equation}
The MSE consists of three terms.
The first two items are \emph{overload distortion} caused by clipping the values of $u$ beyond $[\alpha, \beta]$. The third term means \emph{granular distortion} led by the quantization step $\Delta$. 
For any $u \in [\alpha, \beta]$, its granular distortion is in $[-\frac{\Delta}{2},\frac{\Delta}{2}]$. 
Therefore, it can be reduced by setting an appropriate 
$\Delta$ based on the distribution of $U$. This becomes particularly critical in GNN quantization, as GNNs show large variance at aggregated values~\cite{tailor2020degree}.

\subsection{Challenges with GNN Quantization}\label{gnn_quantization_introduction}
Compared to quantizing CNNs, GNNs involve more types of elements to be quantized with complex interdependencies.
These elements include inputs of each layer, 
 weights,
messages between nodes, inputs and outputs of aggregation stage, and outputs of update stage.
 Since the variance of the updated features after propagation in GNNs is high due to the  varying number of neighbors~\cite{tailor2020degree}, it is particularly challenging to design a low-bit uniform quantizer.  \citet{tailor2020degree} use percentiles to manually decide the quantization range $[\alpha,\beta]$, and a $momentum$ parameter to perform weighted average of the statistics of tensors during training. We empirically observe that the accuracy is highly sensitive to the setting of percentiles and $momentum$, which increases the difficulty of obtaining accurate results especially using low number of bits. \citet{zhu2023rm} learn the quantization step size for each node of input features and each dimension of weights, respectively. However, as it does not quantize the message propagation part, the resulting model size and computations are significantly larger. Moreover, learning parameters per each node yields a higher model parameterization and limits its inductive capabilities including mini-batch training. It is akin to applying $N$ times learned step size~\cite{esser2019learned}, where $N$ is the number of nodes in the graph. 
To address these challenges, we introduce a quantizer with learnable ranges (QLR) which determines the quantization range and 
is also friendly for mini-batch training on large datasets.
\vspace{2pt}

\section{Low-bit quantization for Graph Neural Networks}

This section describes our solution for quantization with learnable ranges (QLR) and a skewness-aware bitwise truncation (BT$^{*}$) that captures the underlying data distribution to preserve accuracy with low-bit representations. 



\subsection{Quantization with Learnable Range}\label{learnable_quantization_range}


GNN involves various components such as layer activations, weights, messages, inputs/outputs of the aggregation and update stages. We aim to quantize all of the aforementioned components to
reduce the model size and maintain high accuracy during inference.

According to the quantization error analysis in Section \ref{quantization_error_analysis},
given quantization level $[\alpha_{q}, \beta_{q}]$, 
$s$ is directly influenced by different settings of $[\alpha, \beta]$. 
Using this observation,  we design a scaling parameter $\gamma$ to modify $[\alpha, \beta]$ into a more data-aware range $[ \gamma\alpha,  \gamma\beta]$,
which updates the quantization range to reduce the quantization error in low-bit cases.



With the learnable quantization range $[ \gamma\alpha,  \gamma\beta]$,
the quantization function can be updated as
\vspace{-2mm}
\begin{equation}
Q(U, s_{\gamma}, z_{\gamma}) = \mathrm{clip}(\lfloor\frac{U}{s_{\gamma}}+z_{\gamma}\rceil, \alpha_q, \beta_q\Big) = U_{q, \gamma}
\label{eq:quantization_function_scale}
\end{equation}
\vspace{-1mm}where $s_{\gamma}$=$\frac{\gamma(\beta-\alpha)}{\beta_{q}-\alpha_{q}}$=${\gamma}s$ is the updated scale, while zero point stays the same $z_{\gamma}$=$\lfloor\frac{\gamma\beta\alpha_q - \gamma\alpha\beta_q}{\gamma\beta-\gamma\alpha}\rceil$ =$z$. The de-quantization function is modified  accordingly
\begin{equation}
D(U_{q,\gamma},  s_{\gamma}, z_{\gamma}) =s_{\gamma}(U_{q, \gamma}-z_{\gamma})=\hat{U}_{\gamma}
\label{eq:dequantization_function_scale}
\end{equation}

To optimize $\gamma$ at the backward propagation, Straight-Through Estimator~\cite{bengio2013estimating} can be used to calculate the gradient of $\gamma$ as  
\begin{equation}
\frac{\partial \hat{U}_{\gamma}}{\partial {\gamma}}=
\begin{cases}
\alpha_{q}, \quad u \leq \alpha_{q}  \\
\beta_{q}, \quad u \geq \beta_{q}  \\
s\lfloor\frac{u}{\gamma s}\rceil - \frac{u}{\gamma}, \quad  \alpha_{q} \textless u \textless \beta_{q}
\end{cases} 
\label{eq:range_boundary_scale}
\end{equation}


QLR learns a scale ($\gamma$) relative to the quantization range of observed values, allocates the limited quantization budget to the remaining observed data points while accounting for the final task. Notably, this is different from learned step size quantization (LSQ)~\cite{esser2019learned}, which optimizes the step size over the full observed values. We have empirically observed that LSQ tends to be highly sensitive to the learning rate. This means that achieving satisfactory accuracy often requires an exhaustive search for the proper hyperparameters. The challenges with LSQ are further amplified because, in GNNs, the value ranges can differ significantly across layers, leading to uneven convergence rates between them.

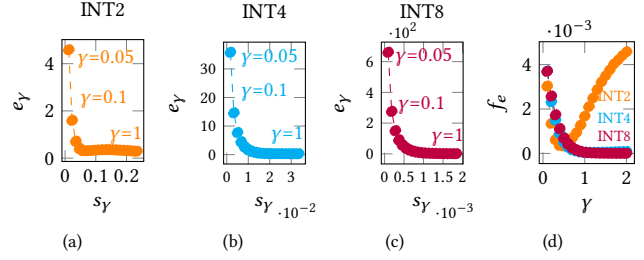
\begin{figure} \vspace{-4mm}
  \centering
 
\centering
 \small
 \hspace{-6mm}
\subfigure[]
 { 
\begin{tikzpicture}
      \begin{axis}
      [width=0.15\textwidth, height=0.18\textwidth, xlabel=$s_{\gamma}$, ylabel={{ $e_{\gamma}$}}, legend to name=legend, title={\small INT2}]
      \addplot [mark=*, orange, dashed]table[x=scale_val,y= error_val]{plots/2_bit_gamma_error.txt};  
        \node[above right, orange] at (axis cs:0.015, 3.5) {$\gamma$=0.05}; 
        \node[above right, orange] at (axis cs:0.025, 1.8) {$\gamma$=0.1};
        \node[above right, orange] at (axis cs:0.12, 0.3) {$\gamma$=1};
      \end{axis}
   \end{tikzpicture}
   }
   \subfigure[]
 { 
  \begin{tikzpicture}
      \begin{axis}
      [width=0.15\textwidth, height=0.18\textwidth, xlabel=$s_{\gamma}$, ylabel={{ $e_{\gamma}$}}, legend to name=legend, title={\small INT4},tick label style={font=\scriptsize}]
      \addplot [mark=*, cyan, dashed]table[x=scale_val,y= error_val]{plots/4_bit_gamma_error.txt};  
        \node[above right, cyan] at (axis cs:0.003, 28) {$\gamma$=0.05}; 
        \node[above right, cyan] at (axis cs:0.003, 16) {$\gamma$=0.1};
        \node[above right, cyan] at (axis cs:0.017, 1) {$\gamma$=1};
      \end{axis} 
      \end{tikzpicture}} 
      \subfigure[]
 { \begin{tikzpicture}  \begin{axis}
      [width=0.15\textwidth, height=0.18\textwidth, xlabel=$s_{\gamma}$, ylabel={{ $e_{\gamma}$}}, legend to name=legend, title={\small INT8},scaled y ticks={base 10:-2},scaled x ticks={base 10:3},tick label style={font=\scriptsize}]
      \addplot [mark=*, purple, dashed]table[x=scale_val,y= error_val]{plots/8_bit_gamma_error.txt};  
        \node[above right, purple] at (axis cs:0.00008, 520) {$\gamma$=0.05}; 
        \node[above right, purple] at (axis cs:0.00015, 210) {$\gamma$=0.1};
        \node[above right, purple] at (axis cs:0.001, 1) {$\gamma$=1};
      \end{axis}
     \end{tikzpicture}}
     \subfigure[]
 { \begin{tikzpicture}
         \begin{axis}
      [width=0.16\textwidth, height=0.18\textwidth, xlabel=${\gamma}$, ylabel={{ $f_e$}}, legend to name=legend]
      \addplot [mark=*, orange, dashed]table[x=alpha_list,y= INT2_error]{plots/overall_gamma_error.txt};  
      \addplot [mark=*, cyan, dashed]table[x=alpha_list,y= INT4_error]{plots/overall_gamma_error.txt}; 
      \addplot [mark=*, purple, dashed]table[x=alpha_list,y= INT8_error]{plots/overall_gamma_error.txt}; 
        \node[above right, orange] at (axis cs:1.13, 0.0020) {\scriptsize INT2};
        \node[above right, cyan] at (axis cs:1.13, 0.0010) {\scriptsize INT4};
        \node[above right, purple] at (axis cs:1.13, 0.0002) {\scriptsize INT8};
      \end{axis}
  \end{tikzpicture}\label{overall_error_plot}}
 \vspace{-2mm}
\caption{Quantization error with different $\gamma$}\label{quantization_error_with_different_gamma}
\vspace{-5pt}
\end{figure}


\begin{figure*}[tb]
\centering
\vspace{-2mm}
\hspace{-6mm}
\vspace{-4mm}
\subfigure[Cora-Message]{
\begin{minipage}[t]{0.25\linewidth}
\centering
\includegraphics[width=1.90in, height=0.9in]{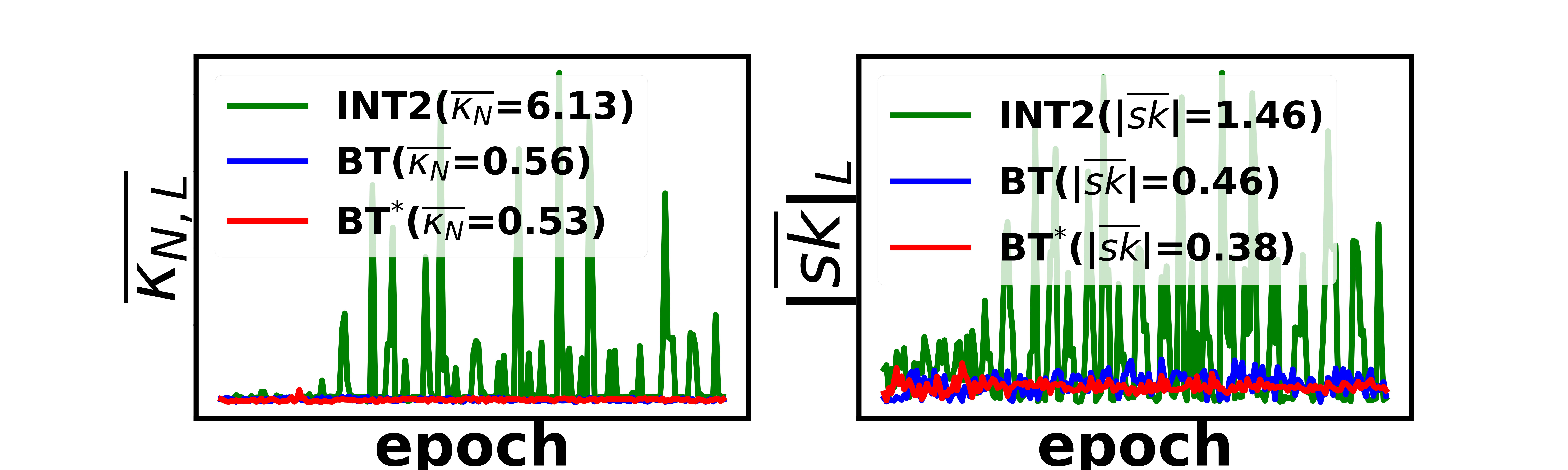}
\end{minipage}
}%
\subfigure[Cora-Aggregate]{
\begin{minipage}[t]{0.25\linewidth}
\centering
\includegraphics[width=1.90in, height=0.9in]{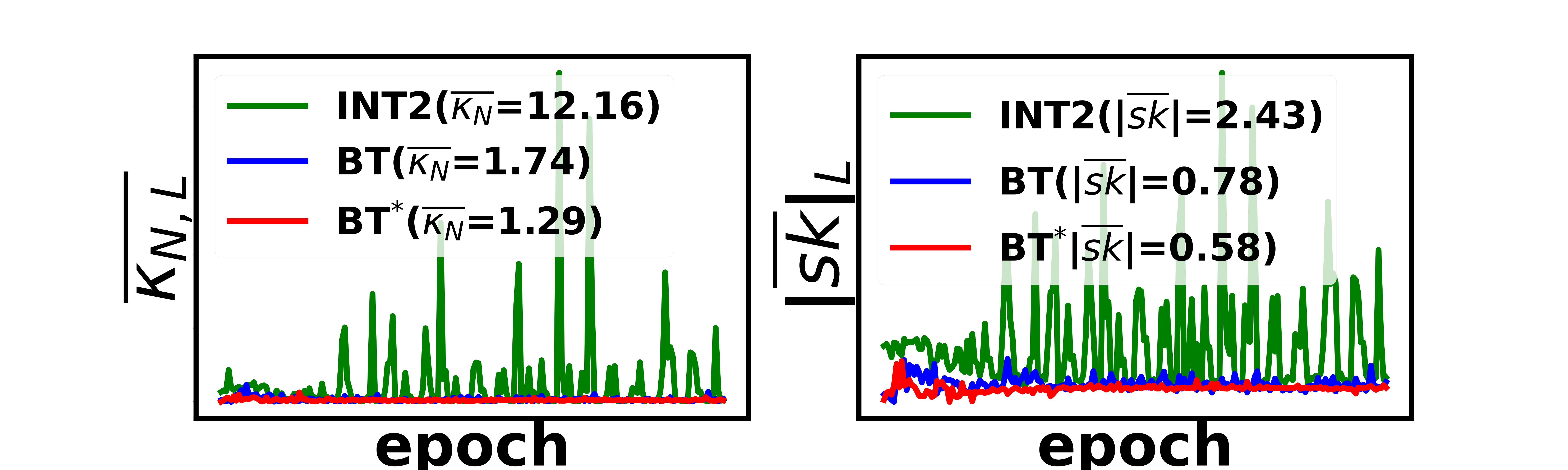
}
\end{minipage}
}%
\subfigure[CS-Message]{
\begin{minipage}[t]{0.25\linewidth}
\centering
\includegraphics[width=1.90in, height=0.9in]{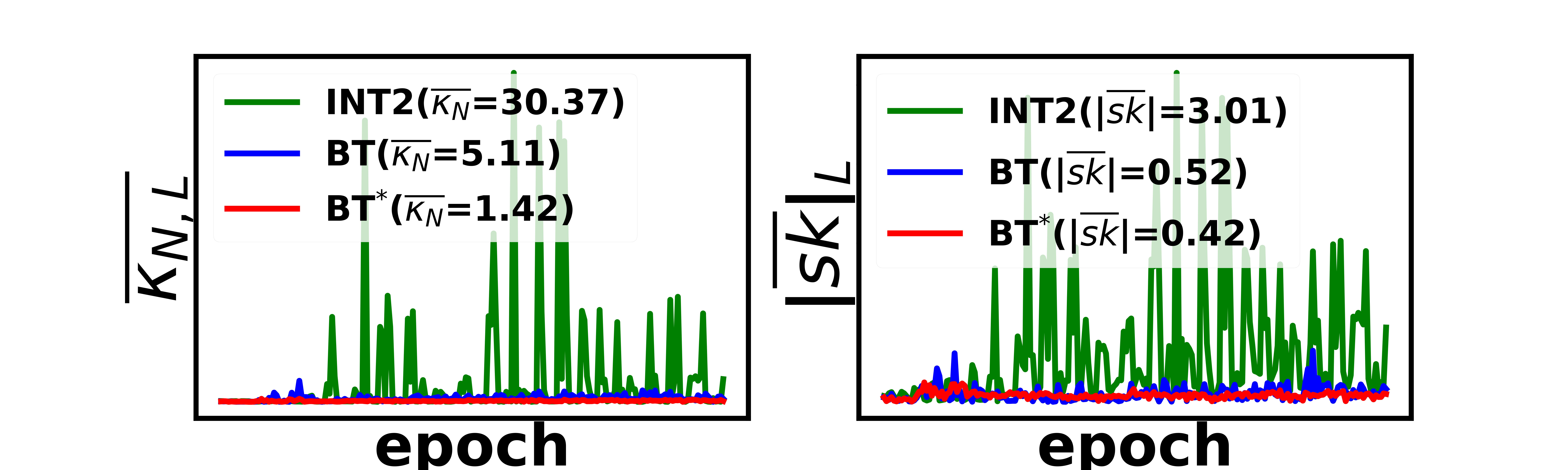}
\end{minipage}
}%
\subfigure[CS-Aggregate]{
\begin{minipage}[t]{0.25\linewidth}
\centering
\includegraphics[width=1.90in, height=0.9in]{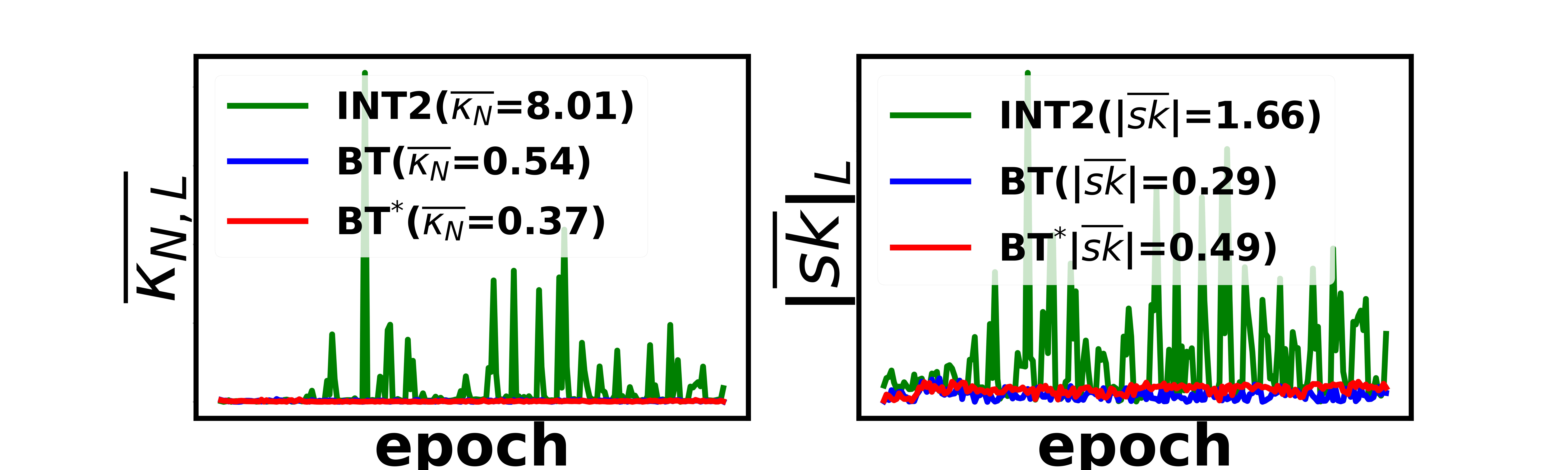}
\end{minipage}
}%
\centering
\caption{Kurtosis and Skewness of different datasets at each epoch }
\label{skewness_kurtosis_epochs_example}
\vspace{-2mm}
\end{figure*}

\noindent\textbf{Quantization error analysis for QLR.} Given a value  $u \in [\alpha, \beta]$, its quantization error can be written as
\begin{equation}
f_{e}(u) = \|u - \hat{u}\|^{2} = \|s_{\gamma}(\frac{u}{s_{\gamma}} - \lfloor\frac{u}{ s_{\gamma}}\rceil)\|^{2}
\label{eq:quantization_error_two_sources}
\end{equation}
It comes from two sources, $s_{\gamma}$ and $\frac{u}{s_{\gamma}} - \lfloor\frac{u}{s_{\gamma}}\rceil$, which are the quantization level and  distortion caused by rounding operation, respectively. 
Figure~\ref{quantization_error_with_different_gamma} shows the distortion error $e_{\gamma}$ and total quantization error $f_e(u)=\|s_{\gamma}e_{\gamma}\|^{2}$, for aggregate output of Cora dataset, across varying scales in $\gamma$$\in$[0.05,1.0].
Noteworthy that, $e_{\gamma}$ and $s_{\gamma}$ affect the error in different directions across different $\gamma$ varying from INT2--INT8. 
Specifically, with an optimized $\gamma$, the distortion in INT2 can be reduced to a scale similar to that of INT4 and INT8 as shown in Figure~\ref{overall_error_plot}. Hence, a learnable $\gamma$ can optimize the quantization range, reducing the total error even in extreme low-bit representations.

\subsection{Skewness-aware Bitwise Truncation}
A basic approach for INT2 can be simply keeping the most (two) significant bits of the higher precision output (e.g., INT8, INT4) as
\begin{equation}
Q_{b_{2}\leftarrow{b_1}}(U_q, s_0) = \lfloor \frac{U_q}{s_0} \rceil s_0
\label{eq:bitwise_truncate}
\end{equation}
where $b_{1}$ and $b_{2}$ ($b_{1}$$\ge$$b_{2}$) are the number of bits used for quantization, $b_{2}$$\leftarrow$${b_1}$ means the $b_{1}$-bit quantized representation being truncated into $b_{2}$ bits; $U_q$ denotes the $b_{1}$-bit representation obtained with Equation \ref{eq:quantization_function_scale}, and $s_0$ is the scale for truncating the low-significant bit representation depending on $b_{1}$ and $b_{2}$. $s_0$ can be obtained as
\begin{equation}
s_0 =  \frac{\alpha_{q_1} - \beta_{q_1}}{\alpha_{q_2} - \beta_{q_2}} 
\label{eq:bitwise_scale}
\end{equation}
where $[\alpha_{q_1}, \beta_{q_1}]$ and $[\alpha_{q_2}, \beta_{q_2}]$ are the quantization levels for $b_1$-bit and $b_2$-bit quantization, respectively. 


While such formulation implicitly assumes the uniformity of $U$, for GNNs this can significantly vary depending on the 
graph topology.
Measures such as kurtosis ($\mathrm{\kappa}$) and skewness ($sk$)~\cite{groeneveld1984measuring} can be employed to better capture information about normality and symmetry of the distribution respectively. 
While prior methods assumed that the neural networks activations follow close-to-normal distributions, we have empirically observed that, for GNNs, these have relatively large kurtosis and are rather asymmetrical on low-bit quantization (Figure~\ref{skewness_kurtosis_epochs_example}). All these bring further challenges in employing bitwise truncation (BT) for GNNs.





We formulate a data-aware truncation mechanism that accounts for skewness of input data. Skewness-aware BT (BT$^{*}$), defined in Equation~\ref{eq:skewness_BT},
can capture the abnormal distribution of quantized elements even under low-bits.
 \vspace{-1pt}
\begin{equation}
{\hat{Q}}_{b_{2}\leftarrow{b_1}}(U_q, s_0) = \lfloor \frac{U_q + \lfloor sk \rceil}{s_0} \rceil s_0
\label{eq:skewness_BT}
\vspace{-1pt}
\end{equation}where $sk$ is the skewness of input tensor $U$.





Figure~\ref{skewness_kurtosis_epochs_example} illustrates the kurtosis and skewness parameters of the message passing and aggregate output blocks for  10-layer SMP, detailed later in Section~\ref{SMP_form_section}, using INT2, INT2-8 (BT) and INT2-8$^{*}$ (BT$^{*}$) quantization across two datasets. We note that kurtosis  ($|\kappa|$) and skewness of the normal distribution are 3 and 0 respectively. Therefore, $\kappa_N= |\kappa-3|$ can be used to measure the normality of the tensor, where smaller $\kappa_N$ means a distribution closer to normal. 
The average kurtosis of BT$^{*}$ remains continuously smaller throughout each epoch, as compared to INT2 and BT, which indicates a robust training process. Similar trends are observed for the skewness $sk$: values for BT$^{*}$ fall in $(-1.0,1.0)$ range. As a result, using skewness ($sk$) in the bitwise truncation process, as provided in Equation~\ref{eq:skewness_BT}, maintains the symmetry of the quantized elements while ensuring their normality. 





\vspace{-6mm}
\section{Layer-wise Smoothness-aware Message Propagation}\label{SMP_form_section}

In GNN learning, each node's feature consists of a true signal, which relates to its class, and a noise component. The essence of message passing is to increase the signal-to-noise ratio by adaptively aggregating node features. However, unexpected or out-of-distribution features from a neighboring node, possibly of different class, can adversely affect the goal of enhancing signal-to-noise ratio. In the asymptotic case, aggregating features from different classes can cause a blending of true features, resulting in oversmoothing. The layer-wise smoothness that preserves locality between layers of GNN, can be helpful in achieving deeper GNNs~\cite{ma2021unified, liu2021elastic}. 

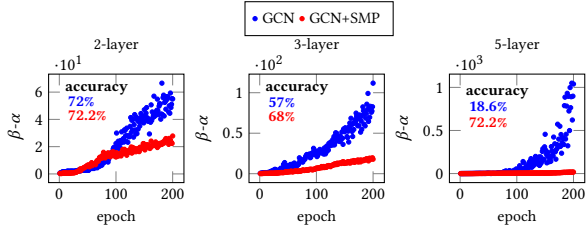
\begin{figure}
\centering
\begin{tikzpicture}
\scriptsize
\matrix{
      \begin{axis}
      [width=0.19\textwidth, height=0.17\textwidth, xlabel=\scriptsize epoch, ylabel={\scriptsize{ $\beta$-$\alpha$}}, legend to name=legend, title={\scriptsize 2-layer},  legend columns=2,scaled y ticks={base 10:-1},]
      \addplot [only marks,blue, mark size=0.8]table[x=X,y= GCN_2]{plots/layer_wise_GCN_quant_range.txt}; 
        \addplot [only marks, red, mark size=0.8]table[x=X,y= GCN_2_SMP]{plots/layer_wise_GCN_quant_range.txt};
        \node[above right] at (axis cs:0, 54) {\scriptsize \textbf{accuracy}}; 
       \node[above right, text=blue] at (axis cs:5, 45) {\scriptsize \textbf{72\%}};
    \node[above right, text=red] at (axis cs:5, 35) {\scriptsize \textbf{72.2\%}};
        
       \legend{ \scriptsize{GCN}, \scriptsize{GCN+SMP}}
      \end{axis}
     
      &
      \begin{axis}
      [width=0.19\textwidth, height=0.17\textwidth, xlabel=\scriptsize epoch, ylabel={\scriptsize{$\beta$-$\alpha$}}, legend to name=legend, title={\scriptsize 3-layer},  legend columns=2,scaled y ticks={base 10:-2},]
      \addplot [only marks,blue, mark size=0.8]table[x=X,y= GCN_3]{plots/layer_wise_GCN_quant_range.txt}; 
        \addplot [only marks, red, mark size=0.8]table[x=X,y=GCN_3_SMP]{plots/layer_wise_GCN_quant_range.txt};
       \node[above right] at (axis cs:0, 90) {\scriptsize \textbf{accuracy}}; 
       \node[above right, text=blue] at (axis cs:5, 75) {\scriptsize \textbf{57\%}};
    \node[above right, text=red] at (axis cs:5, 57) {\scriptsize \textbf{68\%}};
       \legend{ \scriptsize{GCN}, \scriptsize{GCN+SMP}}
      \end{axis} 
      &
      \begin{axis}
      [width=0.19\textwidth, height=0.17\textwidth, xlabel=\scriptsize epoch, ylabel={\scriptsize{$\beta$-$\alpha$}}, legend to name=legend, title={\scriptsize  5-layer},  legend columns=2,scaled y ticks={base 10:-3},]
      \addplot [only marks,blue, mark size=0.8]table[x=X,y= GCN_5]{plots/layer_wise_GCN_quant_range.txt}; 
        \addplot [only marks, red, mark size=0.8]table[x=X,y=GCN_5_SMP]{plots/layer_wise_GCN_quant_range.txt};
       \node[above right] at (axis cs:0, 800) {\scriptsize \textbf{accuracy}}; 
       \node[above right, text=blue] at (axis cs:5, 650) {\scriptsize \textbf{18.6\%}};
    \node[above right, text=red] at (axis cs:5, 470) {\scriptsize \textbf{72.2\%}};
 
       \legend{ \scriptsize{GCN}, \scriptsize{GCN+SMP}}
      \end{axis} 
      \\
   };
    \node at (0.3,1.5) {\ref{legend}};
  \end{tikzpicture}


    

 \vspace{-0.4cm}
\caption{\centering Effect of layer-wise smoothness on the aggregate output quantization range in INT2 quantization}\label{smoothness_GCN_effect}
\vspace{-0.5cm}
\vspace{1pt}

\end{figure}

In Section~\ref{learnable_quantization_range}, we introduced a quantizer that reduces the quantization error by learning an optimal quantization range. We also need to ensure its efficiency with respect to increasing model depth. From our empirical analyses, it is evident that the observed quantization range ($\beta$ -$\alpha$) for low-bit setting expands as the number of layers grows (Figure~\ref{smoothness_GCN_effect}). Based on Equation~\ref{eq:quantized_MSE}, an expanded quantization range directly influences its error. This suggests that quantization may further compromise the accuracy of deeper models which already suffer from over-smoothing. This potential degradation of accuracy is also reflected in Figure~\ref{smoothness_GCN_effect}, where we measure it on GCN with INT2 quantization.

Motivated by the above observations, we devise a layer-wise smoothness approach that brings forth two primary benefits. Firstly, it facilitates smooth message propagation, thereby mitigating the problem of oversmoothing. Secondly, it helps to address the challenges of obtaining satisfactory, low-bit representations caused by substantial and abrupt updates during message propagation.

\noindent\textbf{Outline:} In this section, we present our Smoothness-aware Message Propagation (SMP) solution which aims to reduce the oversmoothing effect and suppress the quantization error to a bound.
We first quantify the layer-wise smoothness and analyze the local smoothness of existing GNNs at message propagation.
We then present the SMP mechanism that smooths the message propagation with a graph denoising approach. After transforming the optimization problem into a Lagrangian function, we develop an optimal solution involving a differential multiplier method (BDMM). 
We also prove the existence of the quantization error bound for quantized SMP.
The results presented in Figure~\ref{smoothness_GCN_effect} (GCN+SMP) confirm that SMP can also help improving the general GCN in INT2 quantization and deeper layer settings.


\subsection{Layer-wise Smoothness}
We quantify the smoothness objective in Definition~\ref{layer_wise_smooth_def} by
measuring the layer-wise local smoothness during message propagation between each GNN layer. 
\vspace{0.5pt}

\begin{definition}(Layer-wise Smoothness)\label{layer_wise_smoothness}
Given a graph $\mathcal{G}$=$(V, E, \mathbf{X})$, the $l$-th layer-wise smoothness is the change of connected nodes $\forall (v_i,v_j)$ $\in$$E$ with a degree normalization from layer $l$-1 to layer $l$.\label{layer_wise_smooth_def}
\end{definition}

The layer-wise smoothness can be formulated as 
\begin{equation}
\centering
\mathbf{S}_{l}=\sum_{(v_{i},v_{j}) \in E}\| (\frac{\mathbf{h}_{i}^{l}}{\sqrt{d_{i}}} -\frac{\mathbf{h}_{j}^{l}}{\sqrt{d_{j}}})- (\frac{\mathbf{h}_{i}^{l-1}}{\sqrt{d_{i}}} -\frac{\mathbf{h}_{j}^{l-1}}{\sqrt{d_{j}}})\|_{2}^{2}
\label{eq:layer_wise_def}
\end{equation}

Specifically, $\mathbf{S}_{l}$ can also be represented as 
\begin{equation}
\centering
\begin{aligned}
\mathbf{S}_{l}&=\sum_{(v_{i},v_{j}) \in E}\| \frac{(\mathbf{h}_{i}^{l}-\mathbf{h}_{i}^{l-1})}{\sqrt{d_{i}}} - \frac{(\mathbf{h}_{j}^{l}-\mathbf{h}_{j}^{l-1})}{\sqrt{d_{j}}}\|_{2}^{2}\\
&=\mathbf{tr((H^{l}-H^{l-1})^{\top}\tilde{L}(H^{l}-H^{l-1}))}
\end{aligned}
\label{eq:layer_wise_transform}
\end{equation}
where $\mathbf{\tilde{L}}$ represents the normalized Laplacian matrix, $\mathbf{\tilde{L}} = \mathbf{I} - \mathbf{\tilde{A}}$, $\mathbf{\tilde{A}}=\mathbf{D}^{-\frac{1}{2}}\hat{\mathbf{A}}\mathbf{D}^{-\frac{1}{2}}$, $\hat{\mathbf{A}} = \mathbf{I} + \mathbf{A}$, and $\mathbf{D_{ii}}=d_{i}=\sum_{j}{\hat{\mathbf{A}}_{ij}}$. The $\mathbf{tr(H^{\top}\tilde{L}H)}$ is the Laplacian regularization to make $\mathbf{H}$ smooth over graph $\mathcal{G}$, similarly, the $l$-th layer-wise smoothness 
$\mathbf{{tr((H^{l}-H^{l-1})^{\top}\tilde{L}(H^{l}-H^{l-1}))}}$ can be explained as smoothing the changes from layer $l$-1 to $l$ over $\mathcal{G}$.

\vspace{3pt}

\subsection{Smoothness-aware Message Propagation}
SMP is designed to guide the training process to achieve local smoothness at message propagation of each layer, by utilizing the smoothness measure presented in Definition~\ref{layer_wise_smooth_def}. Intuitively, SMP aims to avoid drastic correlation/similarity changes for connected nodes to achieve local smoothness at each message update.

We formulate the SMP objective based on graph denoising formulation (at each layer $l \in L$) with degree normalization, 
\begin{equation}
\centering
\thinmuskip=1mu
\medmuskip=2mu
\begin{aligned}
\mathbf{H^{*}} = &\underset{\mathbf{H}}{\text{argmin}} \quad \frac{1}{2}\|{\mathbf{H-X}}\|_{F}^{2} + \frac{\mu}{2} \mathbf{tr(H^{\top}\tilde{L}H)} \\
& \text{subject to: } \mathbf{S}_{l}\leq \delta = \delta_{0}|E|, \forall l \in L \\
\end{aligned}
\label{eq:normalization_denoising}
\end{equation}
where $\mathbf{S}_{l}$=$\mathbf{tr((H-H^{l-1})^{\top}\tilde{L}(H-H^{l-1}))}$,
$\mathbf{\tilde{L}}$ represents the normalized Laplacian matrix, $\mathbf{\tilde{L}} = \mathbf{I} - \mathbf{\tilde{A}}$, $\mathbf{\tilde{A}}=\mathbf{D}^{-\frac{1}{2}}\hat{\mathbf{A}}\mathbf{D}^{-\frac{1}{2}}$, $\hat{\mathbf{A}} = \mathbf{I} + \mathbf{A}$, and $\mathbf{D_{ii}}=d_{i}=\sum_{j}{\hat{\mathbf{A}}_{ij}}$.

The optimization objective aims to find an optimal $\mathbf{H^*}$ which we assume to be the correct feature embedding for the particular graph. We impose three different priors to extract this optimal embedding. The first term minimizes the distance to measure the original feature matrix ($\mathbf{X}$), the second term imposes neighborhood similarity in Equation~\ref{eq:normalization_denoising}. These two objectives have been used in different methods in the literature. In SMP, we impose a new constraint (with $\mathbf{S}_l$) which limits the change of embeddings between layers of the GNN and makes smooth transitions at each message passing iteration. $\delta_{0}$ is the threshold for controlling an allowed variation between the correlations/similarities for the connected node between layers, $|E|$ is the number of edges in the graph.


  
This formulation shows that the constraint aims to mitigate the abrupt changes in the relations between connected nodes due to possibly interfering signals coming from neighboring nodes. 
Alternatively,  $\mathbf{S}_{l}$ can be configured to capture the changes of the smoothness with different distance measures, e.g., $\ell_{1}$ norm.

The Lagrange function for the objective function at layer $l$ has the following form
\begin{equation}
\thinmuskip=1mu
\medmuskip=2mu
 \begin{aligned}\vspace{-0.3cm}
&\mathcal{L}^{l}(\mathbf{H}, \lambda, s) = \underbrace{\frac{1}{2}\|{\mathbf{H-X}}\|_{F}^{2} + \frac{\mu}{2} \mathbf{tr(H^{\top}\tilde{L}H)}}_{\small f(\mathbf{H})} + \lambda g(\mathbf{H},\mathbf{H}^{l-1}, s) \vspace{-0.3cm}
\end{aligned}
\label{eq:formal_lagrangian_form}
 \vspace{-0.2cm}
\end{equation}
where $s$ is slack variable, $\lambda$ is Lagrangian multiplier, and 
 $g(\mathbf{H},\mathbf{H}^{l-1},s)$ = $\delta - \mathbf{S}_{l} - s^{2}$.

Equation~\ref{eq:formal_lagrangian_form} is differentiable with respect to $\mathbf{H}$, $\lambda$ and $s$. However, Lagrangian multiplier method does not directly work with gradient descent optimization and the derivation of optimal solution from KKT (Karush–Kuhn–Tucker) conditions will be cumbersome when the constraints are complex  as in SMP case. Therefore, the optimal solution to Equation~\ref{eq:formal_lagrangian_form} can be derived with the basic differential multiplier method (BDMM)~\cite{platt1988constrained}, which has been proved to optimize Lagrange multipliers in conjunction with the objective argument in a sequential manner.
The BDMM updates are given below
\begin{equation}
\begin{cases}
\mathbf{\overline{\mathbf{H}}}^{l+1} = \mathbf{H}^{l} - \eta_{\mathbf{H}}{\nabla}f(\mathbf{H}^{l})\\
\mathbf{H}^{l+1} = \mathbf{\overline{\mathbf{H}}}^{l+1} -  \mathbf{\eta_{\mathbf{H}}}\lambda^{l}{\nabla}_{\mathbf{\mathbf{H}}}g(\mathbf{\overline{\mathbf{H}}}^{l+1},\mathbf{H}^{l}, s^{l})\\
s^{k+1} = s^{l} + 2 \eta_{s}\lambda^{l}s^{l} \\
\lambda^{k+1} = \lambda^{l} + \eta_{\lambda}g(\mathbf{\mathbf{H}}^{l+1},\mathbf{H}^{l},s^{l+1})\\
\end{cases} 
\label{eq:lagrangian_BBM_update}
\end{equation}

When we calculate the respective gradients in Equation~\ref{eq:formal_lagrangian_form} and incorporate into Equation~\ref{eq:lagrangian_BBM_update}, we easily reach the formulation as
\begin{equation}
\begin{cases}
\mathbf{\overline{\mathbf{H}}}^{l+1} =  (1-(1+\mu)\eta_{\mathbf{H}})\mathbf{\mathbf{H}}^{l} +  \mu\eta_{\mathbf{H}}\tilde{\mathbf{A}}\mathbf{H}^{l} + \eta_{\mathbf{H}}\mathbf{X}\\
\mathbf{H}^{l+1}=\mathbf{\overline{\mathbf{H}}}^{l+1} + 2\eta_{\mathbf{H}}\lambda(\mathbf{I}-\tilde{\mathbf{A}})(\mathbf{\overline{\mathbf{H}}}^{l+1}-\mathbf{H}^{l})\\
s^{l+1} = s^{l} + 2 \eta_{s}\lambda^{l}s^{l} \\
\lambda^{l+1} = \lambda^{l} + \eta_{\lambda}g(\mathbf{H}^{l+1},\mathbf{H}^{l},s^{l+1}) 
\end{cases} 
\label{eq:final_lagrangian_BBM_update}
\end{equation}
where $\eta_{\mathbf{H}}$, $\eta_{\lambda}$, and $\eta_{s}$ are the respective step sizes. 

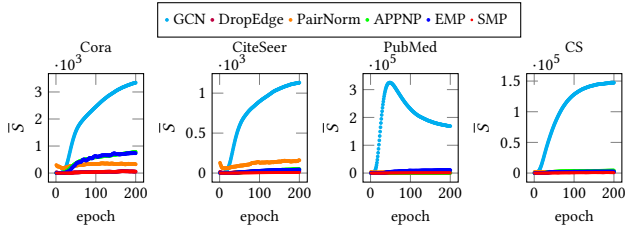
\begin{figure}
\centering
\begin{tikzpicture}
\scriptsize
\matrix{
      \begin{axis}
      [width=0.16\textwidth, height=0.17\textwidth, xlabel=\scriptsize epoch, ylabel={\scriptsize{$\overline{S}$}}, legend to name=legend, title={\scriptsize Cora},  legend columns=6,scaled y ticks={base 10:-3},]
       \addplot [only marks, cyan, mark size=0.5]table[x=x,y= GCN]{plots/epoch_cora_baseline.txt};
      \addplot [only marks,purple, mark size=0.5]table[x=x,y= DropEdge]{plots/epoch_cora_baseline.txt}; 
        \addplot [only marks, orange, mark size=0.5]table[x=x,y= PairNorm]{plots/epoch_cora_baseline.txt};  
        \addplot [only marks,green, mark size=0.5]table[x=x,y= APPNP]{plots/epoch_cora_baseline.txt}; 
        \addplot [only marks, blue, mark size=0.5]table[x=x,y=EMP]{plots/epoch_cora_baseline.txt};
        \addplot [only marks, red, mark size=0.3]table[x=x,y= SMP]{plots/epoch_cora_baseline.txt}; 
       \legend{\scriptsize{GCN}, \scriptsize{DropEdge}, \scriptsize{PairNorm},\scriptsize{APPNP}, 
\scriptsize{EMP},\scriptsize{SMP}, 
 }
      \end{axis}
      &
      \begin{axis}
      [width=0.16\textwidth, height=0.17\textwidth, xlabel=\scriptsize epoch, ylabel={\scriptsize{$\overline{S}$}}, legend to name=legend, title={\scriptsize CiteSeer},  legend columns=6,scaled y ticks={base 10:-3},]
      \addplot [only marks, cyan, mark size=0.5]table[x=x,y= GCN]{plots/epoch_citeseer_baseline.txt};
      \addplot [only marks,purple, mark size=0.5]table[x=x,y= DropEdge]{plots/epoch_citeseer_baseline.txt}; 
        \addplot [only marks, orange, mark size=0.5]table[x=x,y= PairNorm]{plots/epoch_citeseer_baseline.txt};  
        \addplot [only marks,green, mark size=0.5]table[x=x,y= APPNP]{plots/epoch_citeseer_baseline.txt}; 
        \addplot [only marks, blue, mark size=0.5]table[x=x,y=EMP]{plots/epoch_citeseer_baseline.txt};
        \addplot [only marks, red, mark size=0.3]table[x=x,y= SMP]{plots/epoch_citeseer_baseline.txt}; 
       \legend{\scriptsize{GCN}, \scriptsize{DropEdge}, \scriptsize{PairNorm},\scriptsize{APPNP}, 
\scriptsize{EMP},\scriptsize{SMP}, 
 }
      \end{axis}
      &
      \begin{axis}
      [width=0.16\textwidth, height=0.17\textwidth, xlabel=\scriptsize epoch, ylabel={\scriptsize{$\overline{S}$}}, legend to name=legend, title={\scriptsize PubMed},  legend columns=6,scaled y ticks={base 10:-5},]
     \addplot [only marks, cyan, mark size=0.5]table[x=x,y= GCN]{plots/epoch_pubmed_baseline.txt};
      \addplot [only marks,purple, mark size=0.5]table[x=x,y= DropEdge]{plots/epoch_pubmed_baseline.txt}; 
        \addplot [only marks, orange, mark size=0.5]table[x=x,y= PairNorm]{plots/epoch_pubmed_baseline.txt};  
        \addplot [only marks,green, mark size=0.5]table[x=x,y= APPNP]{plots/epoch_pubmed_baseline.txt}; 
        \addplot [only marks, blue, mark size=0.5]table[x=x,y=EMP]{plots/epoch_pubmed_baseline.txt};
        \addplot [only marks, red, mark size=0.3]table[x=x,y= SMP]{plots/epoch_pubmed_baseline.txt}; 
       \legend{\scriptsize{GCN}, \scriptsize{DropEdge}, \scriptsize{PairNorm},\scriptsize{APPNP}, 
\scriptsize{EMP},\scriptsize{SMP}, 
 }
      \end{axis}
      &
      \begin{axis}
      [width=0.16\textwidth, height=0.17\textwidth, xlabel=\scriptsize epoch, ylabel={\scriptsize{$\overline{S}$}}, legend to name=legend, title={\scriptsize CS},  legend columns=6,scaled y ticks={base 10:-5},]
      \addplot [only marks, cyan, mark size=0.5]table[x=x,y= GCN]{plots/epoch_CS_baseline.txt};
      \addplot [only marks,purple, mark size=0.5]table[x=x,y= DropEdge]{plots/epoch_CS_baseline.txt}; 
        \addplot [only marks, orange, mark size=0.5]table[x=x,y= PairNorm]{plots/epoch_CS_baseline.txt};  
        \addplot [only marks,green, mark size=0.5]table[x=x,y= APPNP]{plots/epoch_CS_baseline.txt}; 
        \addplot [only marks, blue, mark size=0.5]table[x=x,y=EMP]{plots/epoch_CS_baseline.txt};
        \addplot [only marks, red, mark size=0.3]table[x=x,y= SMP]{plots/epoch_CS_baseline.txt}; 
       \legend{\scriptsize{GCN}, \scriptsize{DropEdge}, \scriptsize{PairNorm},\scriptsize{APPNP}, 
\scriptsize{EMP},\scriptsize{SMP}, 
 }
      \end{axis}
      \\
    };
    \node at (0.3,1.5) {\ref{legend}};
\vspace{-2mm}
  \end{tikzpicture}
\vspace{-0.7cm}
\caption{\centering Avg layer-wise smoothness against different epochs ($\mathbf{\overline{S}}$) for GNNs}\label{smoothness_obser_example}
\vspace{-11pt}
\end{figure}

\vspace{5pt}\noindent\textbf{Variation of $S_l$ for existing GNNs.} To provide an intuitive understanding of layer-wise smoothness ($\mathbf{S}$), we measure $\mathbf{S}$ for SMP and existing GNNs to quantitatively show how the measure varies with different GNN solutions. We compute $\mathbf{S}$ on 10-layer GCN, SMP, and several existing deep GNN solutions, including DropEdge, APPNP, EMP, and PairNorm. 
Figure~\ref{smoothness_obser_example} shows the average layer-wise smoothness of 10-layer GNNs ($\overline{\mathbf{S}}$) at each epoch, where $\overline{\mathbf{S}}$=$\frac{\sum_{l \in L}\mathbf{S}_{l}}{L-1}$ ($l$$\in$[2,L]).
The GCN, without any safeguard mechanism for oversmoothing, creates extremely large layer-wise variations (higher $\overline{\mathbf{S}}_l$) when compared to deep GNN solutions. 
This example illustrates that deep GNN methods mitigating against oversmoothing are effective to control and increase layer-wise smoothness (decreasing $\overline{\mathbf{S}}_l$) when compared with the general GCN.
Additionally, the $\overline{\mathbf{S}}$ of SMP  drops continuously when compared with other comparable deep GNNs. 
These experiments show that the iterative solution in Equation~\ref{eq:final_lagrangian_BBM_update} is effective in controlling the change between layers by enforcing both node-wise smoothness and layer-wise smoothness. 

\vspace{5pt}
\noindent\textbf{SMP Contribution to Quantization.}
The quantization error for the $l$-th layer representation can be expressed as $f_e^{l}$=$\|\mathbf{H}^{l}-\mathbf{H}^{l,q}\|_2^2$, where $\mathbf{H}^{l,q}$ is the quantized representation of $\mathbf{H}^{l}$.
Accordingly, for the quantized SMP, the smoothness constraint can be written as $S_{l,q}$=$\mathbf{tr((H}^{l,q}-\mathbf{H}^{l-1,q})^{\top}\mathbf{\tilde{L}}\mathbf{(H}^{l,q}-\mathbf{H}^{l-1,q}))$. We can prove $f_{e}^{l}$ is smaller than a bound as provided in LEMMA~\ref{quantization_error_smoothness_proof}, which underlines the superiority of SMP in terms of quantization.
 \begin{customlemma}{1}\label{quantization_error_smoothness_proof} For the $l$-th layer representation $\mathbf{H}^{l}$, 
the quantization error is $f_{e}^{l}
 \leq  l\delta\mathbf{tr}(\mathbf{\Lambda} ^{-1}) + 
  \|\mathbf{H}^{l}- \mathbf{X}^{q}\|_2^2$
    \end{customlemma} 
   \vspace{-3pt}
 \begin{proof}
  The  Laplacian matrix $\tilde{L}$ is eigendecomposable, i.e.,  $\tilde{L}$ = $\mathbf{U\Lambda{U^{\top}}}$, where $\mathbf{U}$ is orthogonal matrix ($\mathbf{U}\mathbf{U}^{\top}$= I ).
   ${\mathbf{S}_{l,q}}$ can be represented as ${\mathbf{S}_{l,q}}$ = $\|\mathbf{\Lambda} ^{\frac{1}{2}}\mathbf{U}^{\top}(\mathbf{H}^{l,q} - \mathbf{H}^{l-1,q})\|_2^2 \leq \delta$. The derivation process is summarized as follows
   \\ 
$\begin{cases}
\|\mathbf{\Lambda}^{\frac{1}{2}}\mathbf{U}^{\top}(\mathbf{H}^{l,q} - \mathbf{H}^{l-1,q})\|_2^{2} = \|(\mathbf{H}^{l,q} - \mathbf{H}^{l-1,q})\mathbf{\Lambda}^{\frac{1}{2}}\|_2^{2}\leq \delta
 \quad\quad\textcircled{1} \\
 \|\mathbf{H}^{l,q} - \mathbf{H}^{l}\|_2^{2} - \|\mathbf{H}^{l-1,q} - \mathbf{H}^{l}\|_2^{2} 
 \leq \|\mathbf{H}^{l,q} - \mathbf{H}^{l-1,q}\|_2^{2} \leq \\
 \|(\mathbf{H}^{l,q} - \mathbf{H}^{l-1,q})\mathbf{\Lambda}^{\frac{1}{2}}\|_2^{2}\|\mathbf{\Lambda}^{-\frac{1}{2}}\|_2^{2} \leq \delta
\mathbf{tr(\Lambda^{-1})}  \quad\quad\quad\quad\quad\quad\quad\textcircled{2}\\
f_e^{l} \leq \delta
\mathbf{tr(\Lambda^{-1})} + \|\mathbf{H}^{l-1,q} - \mathbf{H}^{l}\|_2^{2} \quad \quad \quad \quad \quad \quad\quad \quad \quad  \quad  \quad  \textcircled{3}  \\
\|\mathbf{H}^{i-1,q} - \mathbf{H}^{l}\|_2^{2}
\leq \|\mathbf{H}^{i-1,q} - \mathbf{H}^{i-2,q}\|_2^{2}
+ \|\mathbf{H}^{i-2,q} - \mathbf{H}^{l}\|_2^{2} \\
\quad  \quad = \delta
\mathbf{tr(\Lambda^{-1})} + \|\mathbf{H}^{i-2,q} - \mathbf{H}^{l}\|_2^{2}  $, where $ 
  $1$\leq $i$ \leq l.  \quad \quad  \quad \quad \textcircled{4}\\
f_{e}^{l}
 \leq l\delta\mathbf{tr}(\mathbf{\Lambda} ^{-1}) + 
  \|\mathbf{H}^{l}- \mathbf{X}^{q}\|_2^2 \quad \quad \quad\quad \quad \quad \quad \quad \quad \quad \quad 
 \quad\textcircled{5}\\
\end{cases}$ 
\label{eq:proof_quantize_error_smoothness}
  \end{proof}


 
%

\vspace{-4mm}
\section{Experiments}
This section presents our experiments on benchmark datasets that illustrate the effectiveness of QLR and QLR with BT (BT$^{*}$) quantizers under low-bit settings. 
We also compare our SMP with comparable deep GNN baselines,  
highlighting the capability of SMP in addressing the oversmoothing issue.


\subsection{Experimental Setup}
\label{experiment_setup}
\textbf{Datasets and Baselines.}
Our experiments are performed on five datasets, Cora, PubMed, CiteSeer~\cite{sen2008collective}, CS~\cite{shchur2018pitfalls} and Reddit~\cite{hamilton2017inductive}
in a semi-supervised node classification setting.
The statistics of the datasets are summarized in Table \ref{dataset_statistics}.
\begin{table}[h]
\renewcommand\arraystretch{0.8}\setlength\tabcolsep{3.0pt}
 \caption{Statistics of benchmark datasets 
	}
	\centering
\vspace{-5pt}
\begin{tabular}{cccccc}
\toprule
Dataset  & Cora & CiteSeer & PubMed & CS & Reddit\\
\hline
Nodes     & 2708  & 3327  & 19717     & 18333    &  232965  \\
Edges & 5278  & 4552  & 44324     & 81894     &  114848857 \\
Features   & 1433 & 3703 & 500      & 6805    &   602 \\
Labels       & 7 & 6 & 3     & 15   &   41\\
\hline
\end{tabular}
\vspace{-0.6cm}
 \label{dataset_statistics}
  \vspace{5pt}
\end{table}

\begin{table*}[tb]
\renewcommand\arraystretch{0.8}
\caption{Classification Accuracy ($\%$) of Different Quantization on GCN
	} 
	\small
 \vspace{-3mm}
\centering
 \begin{threeparttable}
\begin{tabular}{c|c|ccc|ccc|ccc}
\hline
Method   &          & \multicolumn{3}{c|}{QLR}                                                       & \multicolumn{3}{c|}{Aggregate-quant}                                     & \multicolumn{3}{c}{Degree-quant}                                                          \\ \hline
Dataset  &     FP       & \multicolumn{1}{c}{INT8}       & \multicolumn{1}{c}{INT4}       & INT2       & \multicolumn{1}{c}{INT8}       & \multicolumn{1}{c}{INT4}       & INT2 & \multicolumn{1}{c}{INT8}                & \multicolumn{1}{c}{INT4}                & INT2 \\ \hline
Cora     & $80.79\pm 1.54$ & \multicolumn{1}{c}{$\textbf{81.03}\pm\textbf{1.03}$} & \multicolumn{1}{c}{$80.40\pm1.06$}  & $74.56\pm2.35$ & \multicolumn{1}{c}{$78.57\pm3.09$} & \multicolumn{1}{c}{$59.03\pm2.50$} & --  & \multicolumn{1}{c}{{$80.64\pm1.47$}} & \multicolumn{1}{c}{{$77.81\pm1.22$}} & --   \\
Citeseer & ${67.74}\pm\textbf{2.29}$ & \multicolumn{1}{c}{$67.59\pm2.07$} & \multicolumn{1}{c}{$67.10\pm2.73$} & $60.12\pm3.02$ & \multicolumn{1}{c}{\textbf{$68.09\pm 1.83$}}         & \multicolumn{1}{c}{$47.79 \pm 4.41$}         & --   & \multicolumn{1}{c}{$67.62\pm2.08$}          & \multicolumn{1}{c}{$66.10\pm1.97$}          & --   \\ 
PubMed   & $78.00\pm2.07$ & \multicolumn{1}{c}{$\textbf{78.74}\pm\textbf{2.56}$} & \multicolumn{1}{c}{$78.10\pm2.58$} & $71.40\pm4.24$ & \multicolumn{1}{c}{$77.60 \pm 2.07$}         & \multicolumn{1}{c}{$68.80 \pm 0.75$}         & $62.26 \pm 0.38$   & \multicolumn{1}{c}{$77.48\pm3.36$}          & \multicolumn{1}{c}{$72.38\pm4.03$}          & --   \\ 
CS       & $90.42\pm0.44$ & \multicolumn{1}{c}{$\textbf{90.51}\pm\textbf{0.45}$} & \multicolumn{1}{c}{$88.42\pm1.62$} & $71.54\pm6.97$ & \multicolumn{1}{c}{$89.45\pm 0.65$}         & \multicolumn{1}{c}{$40.52 \pm 6.74$}         & --   & \multicolumn{1}{c}{$89.96\pm0.58$}          & \multicolumn{1}{c}{$82.51\pm3.09$}          & --   \\ 
Reddit   & $94.04 \pm 0.18$        & \multicolumn{1}{c}{$\textbf{94.13}\pm \textbf{0.07}$}         & \multicolumn{1}{c}{$90.46 \pm 0.08$}         & $85.19 \pm 1.36$       & \multicolumn{1}{c}{$\times$}         & \multicolumn{1}{c}{$\times$}         & $\times$   & \multicolumn{1}{c}{$93.75 \pm 0.20$}                  & \multicolumn{1}{c}{--}                  & --   \\
\hline
\end{tabular}
  \footnotesize
          -- denotes accuracy $\leq$ 40.00$\%$,
          \quad \quad $\times$ indicates the quantization method does not support mini-batch training   

 \end{threeparttable}
\label{GCN_quantization_compare_degreequant}
\vspace{-2mm}
\end{table*}

 \begin{table}[tb]
\renewcommand\arraystretch{0.8}\setlength\tabcolsep{3.0pt}
\caption{\centering Model size (MB) with different number of hidden units ($d$) on different quantization methods
	} 
	\small
\centering
  \vspace{-5pt}
\begin{tabular}{c|c|ccc|ccc|ccc}
\hline
$d$   &          & \multicolumn{3}{c|}{QLR}                                                       & \multicolumn{3}{c|}{ Aggregate-quant}                                     & \multicolumn{3}{c}{ Degree-quant}                                                          \\ \hline
  &     FP       & \multicolumn{1}{c}{INT8}       & \multicolumn{1}{c}{INT4}       & INT2       & \multicolumn{1}{c}{INT8}       & \multicolumn{1}{c}{INT4}       & INT2 & \multicolumn{1}{c}{INT8}                & \multicolumn{1}{c}{INT4}                & INT2 \\ \hline
64     & 1.75& \multicolumn{1}{c}{0.441} & \multicolumn{1}{c}{0.223}  & 0.114 & \multicolumn{1}{c}{3.70} & \multicolumn{1}{c}{3.48} & 3.37   & \multicolumn{1}{c}{0.438} & \multicolumn{1}{c}{0.220} & 0.111   \\ 
128     & 3.49& \multicolumn{1}{c}{{0.878}} & \multicolumn{1}{c}{0.441}  & 0.223& \multicolumn{1}{c}{4.19} & \multicolumn{1}{c}{3.75} & 3.53  & \multicolumn{1}{c}{{0.875}} & \multicolumn{1}{c}{{0.438}} & 0.220  \\ 
512     & 14.0& \multicolumn{1}{c}{3.500} & \multicolumn{1}{c}{1.770}  & 0.878 & \multicolumn{1}{c}{7.79} & \multicolumn{1}{c}{6.05} & 5.17   & \multicolumn{1}{c}{3.490} & \multicolumn{1}{c}{1.750} & 0.875  \\ \hline
\end{tabular}\label{model_size_different_quantization}
\vspace{-5mm}
\end{table}

\begin{table*}[tb]
\renewcommand\arraystretch{0.7}
\setlength\tabcolsep{3.0pt}
 \caption{Classification accuracy of Deep GNN methods (\%) on benchmark datasets 
	}
 \vspace{-3mm}
	\centering

\small
\begin{threeparttable}
\begin{tabular}{c|cccc|cccc|cccc|ccc|ccc}
\hline
                   & \multicolumn{4}{c|}{SMP}                                                                     & \multicolumn{4}{c|}{EMP}                                                                     & \multicolumn{4}{c|}{APPNP}                                                                   & \multicolumn{3}{c|}{PairNorm}                                   & \multicolumn{3}{c}{DropEdge}                                   \\ \hline
Dataset                   & \multicolumn{1}{c}{FP}    & \multicolumn{1}{c}{INT8}  & \multicolumn{1}{c}{INT4}  & INT2  & \multicolumn{1}{c}{FP}    & \multicolumn{1}{c}{INT8}  & \multicolumn{1}{c}{INT4}  & INT2  & \multicolumn{1}{c}{FP}    & \multicolumn{1}{c}{INT8}  & \multicolumn{1}{c}{INT4}  & INT2  & \multicolumn{1}{c}{FP}    & \multicolumn{1}{c}{INT8}  & INT4  & \multicolumn{1}{c}{FP}    & \multicolumn{1}{c}{INT8}  & INT4  \\ \hline
\multirow{2}{*}{Cora}     & \multicolumn{1}{c}{82.91} & \multicolumn{1}{c}{82.92} & \multicolumn{1}{c}{82.69} & 72.69 & \multicolumn{1}{c}{82.59} & \multicolumn{1}{c}{81.58} & \multicolumn{1}{c}{77.93} & 66.20 & \multicolumn{1}{c}{80.78} & \multicolumn{1}{c}{81.82} & \multicolumn{1}{c}{79.68} & 71.28 & \multicolumn{1}{c}{71.20} & \multicolumn{1}{c}{70.21} &  \multirow{2}{*}{--} & \multicolumn{1}{c}{76.08} & \multicolumn{1}{c}{79.10} & 78.31 \\ 
                          & \multicolumn{1}{c}{$\pm 0.64$} & \multicolumn{1}{c}{$\pm 0.51$} & \multicolumn{1}{c}{$\pm0.68$} & $\pm2.48$ & \multicolumn{1}{c}{$\pm0.67$} & \multicolumn{1}{c}{$\pm0.79$} & \multicolumn{1}{c}{$\pm2.62$} & $\pm6.75$ & \multicolumn{1}{c}{$\pm1.42$} & \multicolumn{1}{c}{$\pm1.44$} & \multicolumn{1}{c}{$\pm1.81$} & $\pm2.34$ & \multicolumn{1}{c}{$\pm2.14$} & \multicolumn{1}{c}{$\pm1.50$} &  & \multicolumn{1}{c}{$\pm2.86$} & \multicolumn{1}{c}{$\pm1.00$} & $\pm1.01$ \\ \hline
\multirow{2}{*}{Citeseer} & \multicolumn{1}{c}{71.60} & \multicolumn{1}{c}{71.40}      & \multicolumn{1}{c}{69.76}      &   65.01    & \multicolumn{1}{c}{70.84}      & \multicolumn{1}{c}{71.02}      & \multicolumn{1}{c}{67.53}      &   61.53    & \multicolumn{1}{c}{70.30}      & \multicolumn{1}{c}{68.90}      & \multicolumn{1}{c}{68.77}      &  63.87     & \multicolumn{1}{c}{51.39}      & \multirow{2}{*}{--}     &    \multirow{2}{*}{--}    & \multicolumn{1}{c}{60.07}      & \multicolumn{1}{c}{61.15}      &    59.60   \\ 
                          & \multicolumn{1}{c}{$\pm1.26$} & \multicolumn{1}{c}{$\pm1.58$}      & \multicolumn{1}{c}{$\pm1.63$}      &   $\pm1.26$    & \multicolumn{1}{c}{$\pm1.20$}      & \multicolumn{1}{c}{$\pm1.49$}      & \multicolumn{1}{c}{$\pm2.68$}      &   $\pm3.65$    & \multicolumn{1}{c}{$\pm1.53$}      & \multicolumn{1}{c}{$\pm2.19$}      & \multicolumn{1}{c}{$\pm2.01$}      &  $\pm2.24$     & \multicolumn{1}{c}{$\pm4.16$}      &       &       & \multicolumn{1}{c}{$\pm5.76$}      & \multicolumn{1}{c}{$\pm2.88$}      &  $\pm3.73$     \\ \hline
\multirow{2}{*}{PubMed}   & \multicolumn{1}{c}{79.36} & \multicolumn{1}{c}{79.90}      & \multicolumn{1}{c}{78.44}      &   75.92    & \multicolumn{1}{c}{78.64}      & \multicolumn{1}{c}{78.48}      & \multicolumn{1}{c}{76.18}      &  74.22     & \multicolumn{1}{c}{79.59}      & \multicolumn{1}{c}{79.31}      & \multicolumn{1}{c}{78.49}      &   73.41    & \multicolumn{1}{c}{75.72}      &  \multirow{2}{*}{OOM}      &   \multirow{2}{*}{OOM}     & \multicolumn{1}{c}{75.56}      & \multirow{2}{*}{OOM}       & \multirow{2}{*}{OOM}       \\
   
    & \multicolumn{1}{c}{$\pm2.15$} & \multicolumn{1}{c}{$\pm2.41$}      & \multicolumn{1}{c}{$\pm2.51$}      &   $\pm2.29$    & \multicolumn{1}{c}{$\pm2.93$}      & \multicolumn{1}{c}{$\pm3.32$}      & \multicolumn{1}{c}{$\pm3.44$}      &   $\pm3.03$    & \multicolumn{1}{c}{$\pm1.39$}      & \multicolumn{1}{c}{$\pm1.98$}      & \multicolumn{1}{c}{$\pm2.37$}      &   $\pm1.19 $   & \multicolumn{1}{c}{$\pm2.21$}      & \multicolumn{1}{c}{}      &       & \multicolumn{1}{c}{$\pm1.81$}      & \multicolumn{1}{c}{}      &       \\ \hline
\multirow{2}{*}{CS}       & \multicolumn{1}{c}{92.44} & \multicolumn{1}{c}{92.41}      & \multicolumn{1}{c}{92.24}      &   85.42    & \multicolumn{1}{c}{92.17}      & \multicolumn{1}{c}{91.16}      & \multicolumn{1}{c}{91.81}      & 82.62      & \multicolumn{1}{c}{91.58}      & \multicolumn{1}{c}{92.33}      & \multicolumn{1}{c}{91.94}      &    81.25   & \multicolumn{1}{c}{75.12}      & \multirow{2}{*}{OOM}       &   \multirow{2}{*}{OOM}     & \multicolumn{1}{c}{87.30}      & \multirow{2}{*}{OOM}       &   \multirow{2}{*}{OOM}     \\ 
                          & \multicolumn{1}{c}{$\pm0.64$} & \multicolumn{1}{c}{$\pm0.52$}      & \multicolumn{1}{c}{$\pm0.38$}      &   $\pm 1.80$    & \multicolumn{1}{c}{$\pm0.46$}      & \multicolumn{1}{c}{$\pm0.42$}      & \multicolumn{1}{c}{$\pm$0.54}      &   $\pm$1.09    & \multicolumn{1}{c}{$\pm$0.66}      & \multicolumn{1}{c}{$\pm$0.57}      & \multicolumn{1}{c}{$\pm$0.57}      &  $\pm$8.08     & \multicolumn{1}{c}{$\pm$4.24}      &       &       & \multicolumn{1}{c}{$\pm$1.64}      & \multicolumn{1}{c}{}      &       \\ \hline

\end{tabular} -- denotes accuracy $\leq$ 40.00$\%$, \quad OOM means 'out-of-memory'
\end{threeparttable}
\vspace{-2mm}
\label{allbaselin_quant_performance}
\end{table*}

We start by comparing QLR against two state-of-the-art GNN quantizers, Degree-Quant~\cite{tailor2020degree} and Aggregate-Quant~\cite{zhu2023rm}, on GCN~\cite{kipf2017GCN}. These experiments are complemented with their respective model sizes. To showcase the effectiveness of SMP for quantization and oversmoothing we compare with  4 comparable deep GNN methods, which are, APPNP~\cite{klicpera2018predict}, DropEdge~\cite{rong2019dropedge}, PairNorm~\cite{zhao2019pairnorm} and EMP~\cite{liu2021elastic}. They are evaluated on 10-layer GNNs with 64 hidden units. 
Subsequently, we apply BT (BT$^{*}$) within SMP and EMP pipeline to verify its effectiveness on extreme low-bit representations and its scalability with respect to the numbers of layers.
Finally, we present the runtime efficiency by measuring the throughput under the representations of various quantization levels.


\noindent\textbf{Parameter settings.} For APPNP, DropEdge, PairNorm and EMP, we used the optimal parameters provided within their public repositories. For SMP,  we set the parameters from the following search space: (1) learning rate (lr) $\in$ \{0.005, 0.008, 0.01, 0.015\}; (2) weight decay (wd) $\in$ \{5$e^{-4}$, 1$e^{-4}$, 5$e^{-5}$\}; (3) drop rate $\in$ \{0.8\}; (4) $\mu$ $\in$ \{3, 6, 9\}; (5) the initial value of $\gamma$ $\in$ \{1.0\} ;(6) $\eta_{\lambda}$, $\eta_s$ $\in$ \{1$e^{-5}$,1$e^{-6}$\}; (7) $\delta_{0}$ $\in$ \{0.01, 0.1, 0.5, 1, 2\}. Due to a significant difference in magnitude between the gradients of the scaling parameters ($\gamma$) and other GNN parameters, we have established a distinct search space for the former: learning rate (lr$_{\gamma}$ $\in$ \{0.001, 0.002\})  and weight decay (wd$_{\gamma}$ $\in$ \{1$e^{-4}$, 5$e^{-5}$\}).

 
 We present the average accuracy and standard deviation over 10 random data splits for Cora and CiteSeer, and 5 for PubMed, CS and Reddit. For Reddit dataset, owing to its size, we have employed mini-batch training with a batch$\_$size of 20000. All of the experiments are based on Pytorch~\cite{paszke2019pytorch} and PyTorch Geometric~\cite{fey2019fast}. The experiments are ran on Ubuntu 20.04 with 64GB RAM.
 


\subsection{Comparison with different quantizers}

We compare QLR with the state-of-the-art GNN quantization solutions, Degree-Quant and Aggregate-Quant. Results are summarized in Table \ref{GCN_quantization_compare_degreequant}.

We notice that Aggregate-Quant in default maintains a fixed quantization level of INT4 for weights, while having smaller bits for input features. Moreover, it does not quantize the message-passing blocks of GCN, whereas QLR and Degree-Quant quantize all the elements equally. Hence, for fairness, we also add quantizers for its message-passing blocks and removed the INT4 constraint on its model weights. Also important to note that Aggregate-Quant maintains a step size parameter for each node, which can be viewed as an extension of learned step size quantization (LSQ) \cite{esser2019learned}, which makes it highly inflexible for inductive tasks. Due to that, it does not support mini-batch training, as the topology of the input graph changes with each mini-batch training iteration.

We observe that performance of QLR significantly outperforms those of its competitors irrespective of the quantization level. The approach of optimizing the quantization range in the backward pass makes QLR more robust and effective, especially in low-bit cases. Aggregate-Quant demonstrates superior performance for CiteSeer when applied to INT8 quantization, which can be courtesy of its significantly larger parameter size. 
However, the accuracy of low-bit cases degrades significantly when message passing blocks are also quantized fairly. As for Degree-Quant, while it can achieve comparable performance on INT8, it cannot generate expected performance with INT4 and INT2 quantization on Reddit. Due to its mask sampling strategy and low quantization level, one node sampled to different mini-batches will generate different representations at different batch training, which curbs the overall accurate representation. However, QLR can directly optimize the learnable quantization range based on the observations of subgraphs, hence, it can reduce the comprehensive quantization errors. These further confirm that optimizing the quantization range in QLR enables better preservation of accuracy in low-bit representations.


It is noteworthy that QLR preserves its accuracy results even for INT2 quantization across all datasets, while the alternatives fail to get comparable accuracy. Additionally, QLR even outperforms the full precision (FP) model in INT8 quantization in many cases, showcasing its effectiveness as a noise filter for GNNs.

In Table \ref{model_size_different_quantization}, we report the model sizes of different quantization approaches with varying quantization levels and hidden units ($d$). Due to space limit, we only present the model sizes on CS dataset.
As there is a native 8-bit support, under constant $d$, the sizes of INT8 with QLR and Degree-Quant are consistently reduced to approximately one-fourth of the FP counterpart. For smaller bits, however, we pack INT2 and INT4 similar to the process described in \cite{Vadim:2020:Online}. The size of QLR is slightly larger to that of Degree-quant due to storage of $s_\gamma$, $z_\gamma$ and $\gamma$ parameters. Given the superior accuracy performance of QLR in low-bit settings, its slight increase in model size becomes negligible in comparison.
Overall, with QLR and Degree-quant, the INT2 and INT4 model sizes are significantly smaller than their FP counterparts, with reductions of 16$\times$ and 8$\times$, respectively. However, the size of Aggregate-Quant is 2--6 times that of its counterparts, largely due to 
the dimension and per-node nature of the parameters.

\begin{figure*}[tb]
\centering
\centering
\begin{tikzpicture}
\scriptsize   \matrix{

      \begin{axis}[width=0.28\textwidth, height=0.14\textwidth, xlabel=L, ylabel=Accuracy(\%), legend to name=legend,title={\scriptsize Cora}, legend columns=6
      ]
        \addplot[color=red, mark=otimes*] coordinates{(2,82.1) (4,83.2)(6,83.6)(8,83.4)(10,83.4)(12,83.8)};
        \addplot[color=red, mark=+] coordinates{(2,82.1) (4,82.8)(6,83.2)(8,83.6)(10,83.6)(12,83.5)};
        \addplot[color=blue, mark=otimes*] coordinates{((2,82.8) (4,83.8)(6,83.7)(8,83.6)(10,83.5)(12,84.2)};
        \addplot[color=blue, mark=+] coordinates{((2,82.1) (4,83)(6,83.7)(8,84.1)(10,83.6)(12,83.2)};
        \addplot[color=orange, mark=otimes*] coordinates{(2,82.1) (4,82.2)(6,82.2)(8,83.9)(10,83.4)(12,83.5)};
        \addplot[color=orange, mark=+] coordinates{(2,81) (4,81.5)(6,82)(8,81.5)(10,82.4)(12,80.8)};
        \addplot[color=cyan, mark=otimes*] coordinates{(2,70.1) (4,70.7)(6,73.5)(8,72.2)(10,74.6)(12,73.9)};
        \addplot[color=cyan, mark=+] coordinates{(2,70.6) (4,72.6)(6,71.8)(8,70.9)(10,70)(12,70.7)};
        \addplot[color=magenta, mark=otimes*] coordinates{(2,75) (4,77.8)(6,72.4)(8,77.3)(10,76.1)(12,75.8)};
        \addplot[color=magenta, mark=+] coordinates{((2,71.7) (4,74.2)(6,71.9)(8,73.5)(10,73)(12,73.2)};
        \addplot[color=green, mark=otimes*] coordinates{(2,75.6) (4,78)(6,77.9)(8,77.5)(10,76)(12,77.4)};
        \addplot[color=green, mark=+] coordinates{((2,75.6) (4,76.1)(6,74.7)(8,76.4)(10,76.9)(12,74.9)};
        \legend{ \scriptsize{{SMP-FP}}, \scriptsize{EMP-FP}, \scriptsize{SMP-INT8}, \scriptsize{EMP-INT8},\scriptsize{SMP-INT4}, \scriptsize{EMP-INT4},\scriptsize{SMP-INT2}, \scriptsize{EMP-INT2},\scriptsize{SMP-INT2-8}, \scriptsize{EMP-INT2-8},\scriptsize{SMP-INT2-8{*}}, \scriptsize{EMP-INT2-8{*}}}
      \end{axis} &
      \begin{axis}[width=0.28\textwidth, height=0.14\textwidth, xlabel=L, ylabel=Accuracy(\%),title={\scriptsize CiteSeer}]
        \addplot[color=red, mark=otimes*] coordinates{(2,72.4) (4,72.2)(6,72)(8,71.8)(10,71.5)(12,71.1)};
        \addplot[color=red, mark=+] coordinates{(2,71.7) (4,72)(6,71.8)(8,71.4)(10,71.8)(12,71.5)};
        \addplot[color=blue, mark=otimes*] coordinates{((2,72.8) (4,72.5)(6,72.3)(8,72)(10,72.2)(12,72.4)};
        \addplot[color=blue, mark=+] coordinates{((2,72.5) (4,73.5)(6,72.6)(8,72.7)(10,72.5)(12,73)};
        \addplot[color=orange, mark=otimes*] coordinates{(2,71.7) (4,71.3)(6,71.7)(8,70.7)(10,72.2)(12,71.7)};
        \addplot[color=orange, mark=+] coordinates{(2,69.8) (4,70.8)(6,70.6)(8,70.5)(10,69.9)(12,70.6)};
        \addplot[color=cyan, mark=otimes*] coordinates{(2,65) (4,66.8)(6,68.7)(8,67.8)(10,65.6)(12,65)};
        \addplot[color=cyan, mark=+] coordinates{(2,65.1) (4,64.3)(6,64.8)(8,66.7)(10,66.5)(12,66.3)};
        \addplot[color=magenta, mark=otimes*] coordinates{(2,64.2) (4,66.2)(6,66.1)(8,66.4)(10,69)(12,66.3)};
        \addplot[color=magenta, mark=+] coordinates{(2,63.5) (4,62.2)(6,66.3)(8,63.5)(10,65.3)(12,66.3)};
        \addplot[color=green, mark=otimes*] coordinates{((2,65.9) (4,67.3)(6,70)(8,68.1)(10,68.2)(12,66.2)};
        \addplot[color=green, mark=+] coordinates{((2,63.4) (4,64.8)(6,64.7)(8,65.7)(10,64)(12,63.7)};
      \end{axis}&
      \begin{axis}[width=0.28\textwidth, height=0.14\textwidth, xlabel=\scriptsize{L}, ylabel=Accuracy(\%),title={\scriptsize{PubMed}}]
        \addplot[color=red, mark=otimes*] coordinates{(2,81) (4,80.8)(6,80.7)(8,81)(10,80.7)(12,81)};
        \addplot[color=red, mark=+] coordinates{(2,81.2) (4,81)(6,80.6)(8,80.6)(10,80.5)(12,80.7)};
        \addplot[color=blue, mark=otimes*] coordinates{((2,76.8) (4,80.2)(6,80.8)(8,81)(10,82.2)(12,81)};
        \addplot[color=blue, mark=+] coordinates{((2,81.2) (4,81.3)(6,80.8)(8,81.3)(10,81.5)(12,81.4)};
        \addplot[color=orange, mark=otimes*] coordinates{(2,76) (4,79)(6,80.1)(8,79.8)(10,81.1)(12,79.8)};
        \addplot[color=orange, mark=+] coordinates{(2,78.4) (4,79.6)(6,79.5)(8,80)(10,80.2)(12,79.4)};
        \addplot[color=cyan, mark=otimes*] coordinates{(2,74.6) (4,76.3)(6,78.5)(8,77.5)(10,77.4)(12,77)};
        \addplot[color=cyan, mark=+] coordinates{(2,74.7) (4,77.2)(6,76.6)(8,76.5)(10,76.3)(12,75.6)};
        \addplot[color=magenta, mark=otimes*] coordinates{(2,73.8) (4,75.2)(6,76.7)(8,76.2)(10,77.7)(12,75.4)};
        \addplot[color=magenta, mark=+] coordinates{(2,77.2) (4,75.7)(6,75.4)(8,73.4)(10,73.7)(12,74)};
        \addplot[color=green, mark=otimes*] coordinates{(2,76.9) (4,78.8)(6,78)(8,77.2)(10,78.2)(12,77)};
        \addplot[color=green, mark=+] coordinates{((2,77.9) (4,78)(6,78.2)(8,76.2)(10,75.4)(12,76.5)};
      \end{axis} &
      \begin{axis}[width=0.28\textwidth, height=0.14\textwidth, xlabel=\scriptsize{L}, ylabel=Accuracy(\%),title={\scriptsize{CS}}]
        \addplot[color=red, mark=otimes*] coordinates{(2,91.81) (4,91.82)(6,91.16)(8,90.67)(10,90.39)(12,90.99)};
        \addplot[color=red, mark=+] coordinates{(2,91.81) (4,91.88)(6,91.32)(8,91.38)(10,91.66)(12,91.86)};
        \addplot[color=blue, mark=otimes*] coordinates{((2,91.67) (4,91.73)(6,92.01)(8,91.91)(10,92.02)(12,92.08)};
        \addplot[color=blue, mark=+] coordinates{((2,92.05) (4,91.83)(6,91.84)(8,91.98)(10,91.99)(12,91.86)};
        \addplot[color=orange, mark=otimes*] coordinates{(2,90.62) (4,91.07)(6,91.22)(8,91.12)(10,91.48)(12,90.96)};
        \addplot[color=orange, mark=+] coordinates{(2,90.84) (4,90.54)(6,90.91)(8,91.13)(10,90.03)(12,90.5)};
        \addplot[color=cyan, mark=otimes*] coordinates{(2,85.57) (4,88.3)(6,88.27)(8,86.92)(10,86.21)(12,90.01)};
        \addplot[color=cyan, mark=+] coordinates{(2,87.99) (4,80.27)(6,80.53)(8,77.27)(10,77.27)(12,86.05)};
        \addplot[color=magenta, mark=otimes*] coordinates{(2,89.46) (4,90.85)(6,90.51)(8,90.66)(10,90.56)(12,90.43)};
        \addplot[color=magenta, mark=+] coordinates{(2,85.25) (4,85.43)(6,82.8)(8,85.32)(10,85.09)(12,83.81)};
        \addplot[color=green, mark=otimes*] coordinates{(2,89.68) (4,90.39)(6,90.95)(8,90.29)(10,90.89)(12,90.53)};
        \addplot[color=green, mark=+] coordinates{(2,88.32) (4,87.57)(6,91.12)(8,91.89)(10,91.11)(12,87.24)};
      \end{axis}\\
    };
    \node at (0.2,1.5) {\ref{legend}
                };

  \end{tikzpicture}
\centering
\vspace{-0.5cm}
\caption{Results of SMP and EMP with varying layers}\label{different_layer_results}
\label{different_layer_results}
\vspace{-2mm}
\end{figure*}
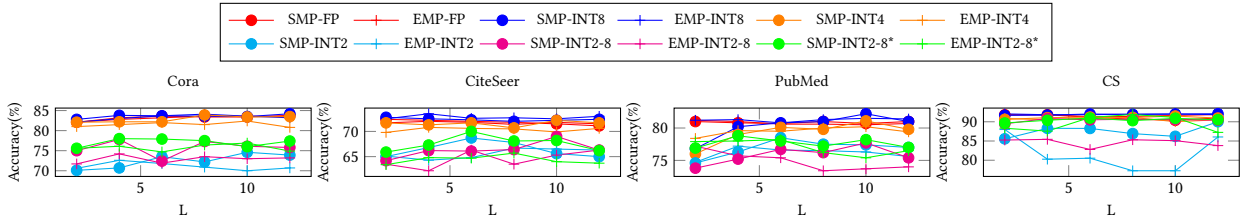
%

\vspace{-2pt}
\subsection{Comparisons with existing deep GNNs} \label{deep_architecture_quant}
\subsubsection{Node classification with existing deep GNNs}
We compare SMP with the existing deep GNNs in terms of both full-precision (FP) and quantized models using QLR. Table \ref{allbaselin_quant_performance} presents the classification accuracy results using 10-layer GNN. Notably,  SMP consistently outperforms the alternative methods on Cora, CiteSeer and CS with FP models, and slightly lower than that of APPNP on PubMed. SMP improves over EMP by enforcing smoothness at layer-wise message propagation during training and inference. 

For quantized models, although INT8 achieves high accuracy close to FP for all methods, INT4 performance of DropEdge and PairNorm drops significantly, rendering it incomparable in some cases, and throws OOM errors on larger datasets. This is primarily due to the complexity of the "backbone" models \cite{rong2019dropedge}. For example, PairNorm, employing a GCN backbone, trains a weight matrix for each layer. Likewise, DropEdge utilizes more intricate backbones such as GCN, ResGCN \cite{he2016deep}, IncepGCN \cite{szegedy2016rethinking}, and introduces connection perturbations at every layer. Consequently, these factors contribute to higher computational and storage requirements, which further escalate with the depth of the model. On the contrary, EMP, APPNP, and SMP employ much simpler architectures that involve only two weight matrices prior to GNN propagation. As a result, these models have more moderate and scalable requirements, making them more amenable for deep GNN quantization.

SMP also consistently improves the low-bit performance enabling more stable training as compared to other methods. This can be explained by the smooth, narrow-ranged representations across layers, enabling QLR to identify more precise low-bit representations. The empirical results are also in line with the quantization error upper bound for quantized SMP as proved in LEMMA \ref{quantization_error_smoothness_proof}. We also note that QLR can also improve EMP and APPNP to achieve reasonable accuracy with graceful degradation even in INT2 quantization. This highlights the importance of optimizing the quantization range for informative representation.


 \vspace{-2pt}
\subsubsection{Performance with varying number of GNN layers} Figure \ref{different_layer_results} presents the performance of SMP and EMP with respect to different number of GNN layers, using FP, INT8, INT4, INT2, INT2-8 ( BT), and INT2-8$^{*}$( BT$^{*}$) representations. For simplicity, we narrow the search space as lr $\in$ \{0.005, 0.008, 0.01, 0.015\}, wd=5$e^{-4}$, lr$_\gamma$=5$e^{-4}$, wd$_\gamma$=1$e^{-5}$. We note that SMP-FP and SMP-INT8 outperform EMP-FP and EMP-INT8 in most cases with varying margins. 
The improvements can be further enhanced by tuning the parameters in a wider search space as listed in Section \ref{experiment_setup}.
We note that SMP-INT4 continuously outperforms EMP-INT4 in nearly all cases with different values of $L$ (except $L$=2--4 on PubMed and $L$=2 on CS). SMP-INT2 achieves relatively high accuracy when compared with EMP-INT2, which underlines the benefits of smoothness constraint of SMP for extreme low-bit quantization.

\vspace{-2pt}
\subsubsection{Effect of Bitwise Truncation on GNN quantization}
For INT2 quantization in Figure \ref{different_layer_results}, the performance of BT (INT2-8) and BT$^{*}$ (INT2-8$^{*}$) outperforms INT2 quantization with the basic QLR in most cases, 
while the accuracy of EMP-INT2-8$^{*}$ on CiteSeer is lower (around 0.1$\%$--2.5$\%$) than that of INT2 at $L$=6--12. Similar results are observed with PubMed at SMP-INT2-8$^{*}$ when $L$=6--8 with margins of 0.1$\%$--1.3$\%$. When we compare INT2-8 and INT2-8$^{*}$ under the same circumstances, the accuracy of INT2-8$^{*}$ is significantly larger than that of INT2-8 in most cases (except for CiteSeer, the performance of INT2-8$^{*}$ is slightly smaller than that of INT2-8 on SMP at $L$=6--8 and EMP at $L$=8). 
The performance of INT2-8$^{*}$ demonstrates the advantage over INT2-8 on the  large datasets, i.e., PubMed and CS, and the accuracy of INT2-8$^{*}$ is close to that of INT4. Especially, the accuracy of INT2-8$^{*}$ is, continuously, around 5$\%$--7$\%$ higher than that of INT2-8 on CS dataset with respect to varying number of layers.  This shows the effectiveness of BT$^{*}$ in reaching relatively high accuracy with low-bit representations.


\begin{figure}
\vspace{-0.3cm}
\centering
\begin{tikzpicture}
\definecolor{color1}{HTML}{ff0000} 
\definecolor{color2}{HTML}{00ff00} 
\definecolor{color3}{HTML}{0000ff} 
\definecolor{color4}{HTML}{ffff00} 
\definecolor{color5}{HTML}{00ffff} 
\definecolor{color6}{HTML}{ff00ff} 
\definecolor{color7}{HTML}{ffa500} 
\definecolor{color8}{HTML}{800080} 
\definecolor{color9}{HTML}{008080} 
\definecolor{color10}{HTML}{ff69b4} 
\definecolor{color11}{HTML}{008000} 
\definecolor{color12}{HTML}{000080} 
\definecolor{color13}{HTML}{800000} 
\definecolor{color14}{HTML}{ffa07a} 
\definecolor{color15}{HTML}{f08080} 
\definecolor{color16}{HTML}{bdb76b} 
\definecolor{color17}{HTML}{00ff7f} 
\definecolor{color18}{HTML}{ff1493} 
\definecolor{color19}{HTML}{7fffd4} 
\definecolor{color20}{HTML}{9370db} 


\matrix{
      \begin{axis}[width=0.175\textwidth, height=0.18\textwidth, xlabel=\scriptsize epoch, ylabel=\scriptsize{Accuracy (\%)}, legend to name=legend,title={\scriptsize Train Accuracy}, legend columns=5,,legend style={minimum size=0.05cm},]
       \addplot [only marks, color1, mark size=0.4]table[x=x,y=INT8_SMP_train]{plots/torch_train_list.txt}; 
        \addplot [only marks, color2, mark size=0.4]table[x=x,y=INT8_EMP_train]{plots/torch_train_list.txt};
       \addplot [only marks, color3, mark size=0.4]table[x=x,y=INT4_SMP_train]{plots/torch_train_list.txt}; 
        \addplot [only marks, black, mark size=0.4]table[x=x,y=INT4_EMP_train]{plots/torch_train_list.txt};
        \addplot [only marks, color5, mark size=0.4]table[x=x,y=INT2_SMP_train_acc]{plots/torch_train_list.txt}; 
        \addplot [only marks, color6, mark size=0.4]table[x=x,y=INT2_EMP_train_acc]{plots/torch_train_list.txt};
        \addplot [only marks, color7, mark size=0.4]table[x=x,y=INT1_no_skew_SMP_train_acc]{plots/torch_train_list.txt}; 
        \addplot [only marks, color8, mark size=0.4]table[x=x,y=INT1_no_skew_EMP_train_acc]{plots/torch_train_list.txt};
        \addplot [only marks, color9, mark size=0.4]table[x=x,y=INT1_skew_SMP_train_acc]{plots/torch_train_list.txt}; 
    \addplot [only marks, color10, mark size=0.4]table[x=x,y=INT1_skew_EMP_train_acc]{plots/torch_train_list.txt}; 

        \legend{ \scriptsize{SMP-INT8}, \scriptsize{EMP-INT8}, \scriptsize{SMP-INT4}, \scriptsize{EMP-INT4}, \scriptsize{SMP-INT2},\scriptsize{EMP-INT2}, \scriptsize{SMP-INT8-2}, \scriptsize{EMP-INT8-2}, \scriptsize{SMP-INT8-2*}, \scriptsize{EMP-INT8-2*}}
      \end{axis} &
      \begin{axis}[width=0.175\textwidth, height=0.18\textwidth, xlabel= \scriptsize epoch, ylabel=\scriptsize{Accuracy (\%)},title={\scriptsize Valid Accuracy}]
     \addplot [only marks, color1, mark size=0.4]table[x=x,y=INT8_SMP_valid]{plots/torch_valid_list.txt}; 
        \addplot [only marks,color2, mark size=0.4]table[x=x,y=INT8_EMP_valid]{plots/torch_valid_list.txt};
       \addplot [only marks, color3, mark size=0.4]table[x=x,y=INT4_SMP_valid]{plots/torch_valid_list.txt}; 
        \addplot [only marks, black, mark size=0.4]table[x=x,y=INT4_EMP_valid]{plots/torch_valid_list.txt};
        \addplot [only marks, color5, mark size=0.4]table[x=x,y=INT2_SMP_valid_acc]{plots/torch_valid_list.txt}; 
        \addplot [only marks,color6, mark size=0.4]table[x=x,y=INT2_EMP_valid_acc]{plots/torch_valid_list.txt};
        \addplot [only marks, cyan, color7, mark size=0.4]table[x=x,y=INT1_no_skew_SMP_valid_acc]{plots/torch_valid_list.txt}; 
        \addplot [only marks, color8, mark size=0.4]table[x=x,y=INT1_no_skew_EMP_valid_acc]{plots/torch_valid_list.txt};
        \addplot [only marks,color9, mark size=0.4]table[x=x,y=INT1_skew_SMP_valid_acc]{plots/torch_valid_list.txt}; 
    \addplot [only marks, color10, mark size=0.4]table[x=x,y=INT1_skew_EMP_valid_acc]{plots/torch_valid_list.txt}; 
      \end{axis}&
      \begin{axis}[width=0.175\textwidth, height=0.18\textwidth, xlabel= \scriptsize epoch, ylabel=\scriptsize{Accuracy (\%)},title={\scriptsize Test Accuracy}]
       \addplot [only marks, color1, mark size=0.4]table[x=x,y=INT8_SMP_test]{plots/torch_test_list.txt}; 
        \addplot [only marks, color2, mark size=0.4]table[x=x,y=INT8_EMP_test]{plots/torch_test_list.txt};
       \addplot [only marks, color3, mark size=0.4]table[x=x,y=INT4_SMP_test]{plots/torch_test_list.txt}; 
        \addplot [only marks, black, mark size=0.4]table[x=x,y=INT4_EMP_test]{plots/torch_test_list.txt};
        \addplot [only marks, color5, mark size=0.4]table[x=x,y=INT2_SMP_test_acc]{plots/torch_test_list.txt}; 
        \addplot [only marks, color6, mark size=0.4]table[x=x,y=INT2_EMP_test_acc]{plots/torch_test_list.txt};
        \addplot [only marks, color7, mark size=0.4]table[x=x,y=INT1_no_skew_SMP_test_acc]{plots/torch_test_list.txt}; 
        \addplot [only marks, color8, mark size=0.4]table[x=x,y=INT1_no_skew_EMP_test_acc]{plots/torch_test_list.txt};
        \addplot [only marks, color9, mark size=0.4]table[x=x,y=INT1_skew_SMP_test_acc]{plots/torch_test_list.txt}; 
    \addplot [only marks, color10, mark size=0.4]table[x=x,y=INT1_skew_EMP_test_acc]{plots/torch_test_list.txt}; 
      \end{axis}\\
    };
    \node at (0.3,2.0){\ref{legend}};
\vspace{-2mm}

  \end{tikzpicture}
\vspace{-0.9cm}
\caption{\centering Quantization performance of SMP and EMP on CS }\label{training_loss_plot}
\vspace{-4mm}

\end{figure}
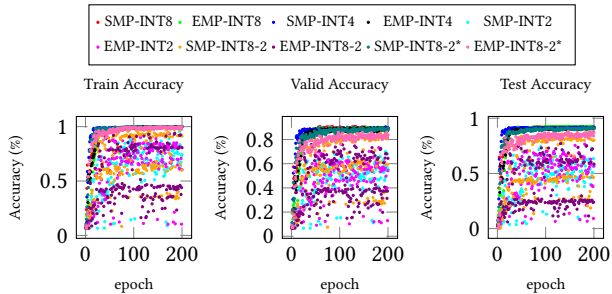


Figure \ref{training_loss_plot} presents the full training process of SMP and EMP with variations of our quantization approach on CS dataset.
While SMP and EMP show unstable performance with INT2 representation generated by the basic quantizer, applying skewness-aware BT (BT$^{*}$) contributes improvements that are close to that of INT8 quantization. Furthermore, SMP is more robust than EMP, due to the existence of quantization error bound in SMP.

\vspace{-3pt}
\subsection{Inference Speedup}
In Table \ref{inference_speedup}, we elucidate the inference times associated with varying quantization levels for the 2-layer GCN and SMP architectures, respectively. These model inferences are conducted on the Reddit dataset. The $\uparrow$ signifies the inference speedup comparing with FP model. To realize quantized GNN speed improvements across different quantization levels, we leverage the recent Tensor Core-based approach, QGCT \cite{wang2022qgtc}, applied to both GCN and SMP. We observe a notable speedup of 5.11 $\times$ and 6.44 $\times$ with SMP and GCN, respectively, in the context of low-bit representation (INT2), in comparison to the FP counterparts. Notably, the speedup for SMP exhibits a slight reduction compared to GCN, attributed to the additional computational overhead of SMP. Remarkably, with the same number of layer ($L$), SMP showcases superior accuracy performance relative to GCN. This is exemplified in Table \ref{GCN_quantization_compare_degreequant} and Figure \ref{different_layer_results} in the CS dataset with $L$=2. Specifically, for SMP, the INT8 and INT4 accuracy outperforms GCN by approximately 2$\%$, while SMP in INT2 mode demonstrates a performance advantage over GCN by up to 13.5$\%$.
 \begin{table}[tb]
\renewcommand\arraystretch{0.9}\setlength\tabcolsep{3.0pt}
\caption{\centering Inference time (ms) of SMP and GCN in different quantization levels on Reddit dataset
	} 
	\small
\centering
  \vspace{-5pt}
  
\begin{tabular}{c|c|cc|cc|cc}
\hline
    & FP     & INT8  & $\uparrow$ & INT4  & $\uparrow$ & INT2  & $\uparrow$ \\ \hline
SMP & 178.3  & 46.46 & 3.84 $\times$ & 37.96 & 4.70 $\times$ & 34.89 & 5.11 $\times$ \\ \hline
GCN & 156.97 & 34.96 & 4.49 $\times$ & 27.29 & 5.75 $\times$ & 24.37 & 6.44 $\times$ \\ \hline
\end{tabular}\label{inference_speedup}
\vspace{-9pt}
\end{table}



\vspace{-2mm}
\section{Conclusion}
We have introduced an end-to-end solution towards achieving scalable deep GNNs, involving an efficient quantization with learnable ranges, with skewness-aware bitwise truncation, and a smoothness-aware message propagation (SMP) mechanism for efficient training and managing large deep GNNs. The solution reduces the model size and maintains its accuracy for classification even in low-bit representations.  The message passing block in training is enforced to have layer-wise smoothness and constrains the changes between neighbor nodes. We formulate it as an additional constraint to a graph denoising optimization function and solved by Lagrange functions with an iterative BDMM algorithm. It aims to mitigate the  oversmoothing problem in GNNs and to avoid the performance degradation encountered in low-bit quantization-aware training. We provide an upper bound on the error for the quantized SMP algorithm. Experiments show how the proposed solution achieves significant improvements over the-state-of-the-art approaches, providing a significant reduction in model sizes, an order of magnitude smaller than the full precision (FP) model with comparable accuracy results, and mitigating the oversmoothing problem on benchmark datasets.



\begin{acks}
This work is supported in part by the UK Engineering and Physical Sciences Research Council under Grant No. EP/T51794X/1.
\end{acks}



 \bibliographystyle{ACM-Reference-Format}
 \balance
\bibliography{main}
\end{document}